\documentclass[journal]{IEEEtran}
\usepackage[ruled,linesnumbered]{algorithm2e}
\usepackage{algpseudocode}
\usepackage{listings}

\usepackage{float}
\usepackage{graphicx}
\usepackage{multirow}
\usepackage{amsmath} 
\usepackage{amssymb}  
\usepackage[normalem]{ulem}
\usepackage{booktabs}
\usepackage{footnote}
\usepackage{threeparttable}
\usepackage[colorlinks,linkcolor=red]{hyperref}
\usepackage{cite}
\usepackage{color}
\usepackage{booktabs}
\usepackage[table]{xcolor}
\usepackage{caption}
\usepackage{float} 
\usepackage{subcaption}

\usepackage{xcolor}
\definecolor{mycolorpurple}{RGB}{255, 102, 255}
\definecolor{mycolorred}{RGB}{227, 26, 26}
\definecolor{mycolorskyblue}{RGB}{6, 185, 238}

\begin{document}

\title{\textcolor{black}{VoxelNextFusion}: A Simple, Unified and Effective Voxel Fusion Framework for Multi-Modal 3D Object Detection}

\author{Ziying Song, Guoxin Zhang, Jun Xie,  Lin Liu, Caiyan Jia, Shaoqing Xu, Zhepeng Wang 
\thanks{This work was supported in part by the National Key R\&D Program of China (2018AAA0100302), supported by the STI 2030-Major Projects under Grant 2021ZD0201404.\emph{(Corresponding author: Caiyan Jia.)}}
\thanks{Ziying Song, Lin Liu, Caiyan Jia are with School of Computer and Information Technology, Beijing Key Lab of Traffic Data Analysis and Mining, Beijing Jiaotong University, Beijing 100044, China (e-mail: 22110110@bjtu.edu.cn; liulin010811@gmail.com; cyjia@bjtu.edu.cn.)}

\thanks{Guoxin Zhang is with Hebei University of Science and Technology, School of Information Science and Engineering,  Shijiazhuang 050018, China, and also work done during an internship at Lenovo Research, Beijing 100085, China. (e-mail: zhangguoxincs@gmail.com).}

\thanks{Jun Xie,  Zhepeng Wang are with Lenovo Research, Beijing 100085, China (xiejun@lenovo.com, wangzpb@lenovo.com).}

\thanks{Shaoqing Xu is with the State Key Laboratory of Internet of Things for Smart City and Department of Electrome chanical Engineering, University of Macau, Macau 999078, China (e-mail: shaoqing.xu@connect.um.edu.mo)
}


}

\maketitle

\begin{figure*}[t]
    \centering
    \captionsetup[subfloat]{font=normalsize}
    \subfloat[Illustration of resolution mismatch]
    {\includegraphics[height=2in,width=2.5in]{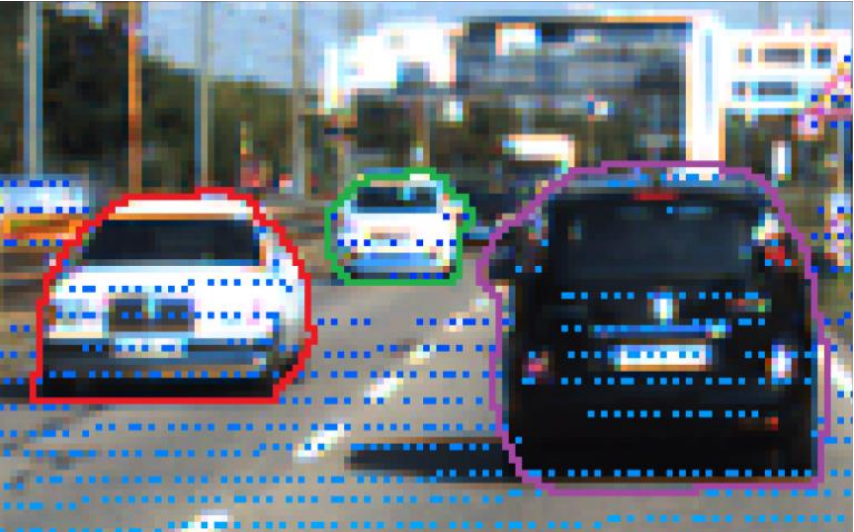}%
    \label{fig_first_case}
    }
    \hfil
    \subfloat[\textcolor{black}{VoxelNextFusion} Strategy]
    {\includegraphics[height=2in,width=1.91in]{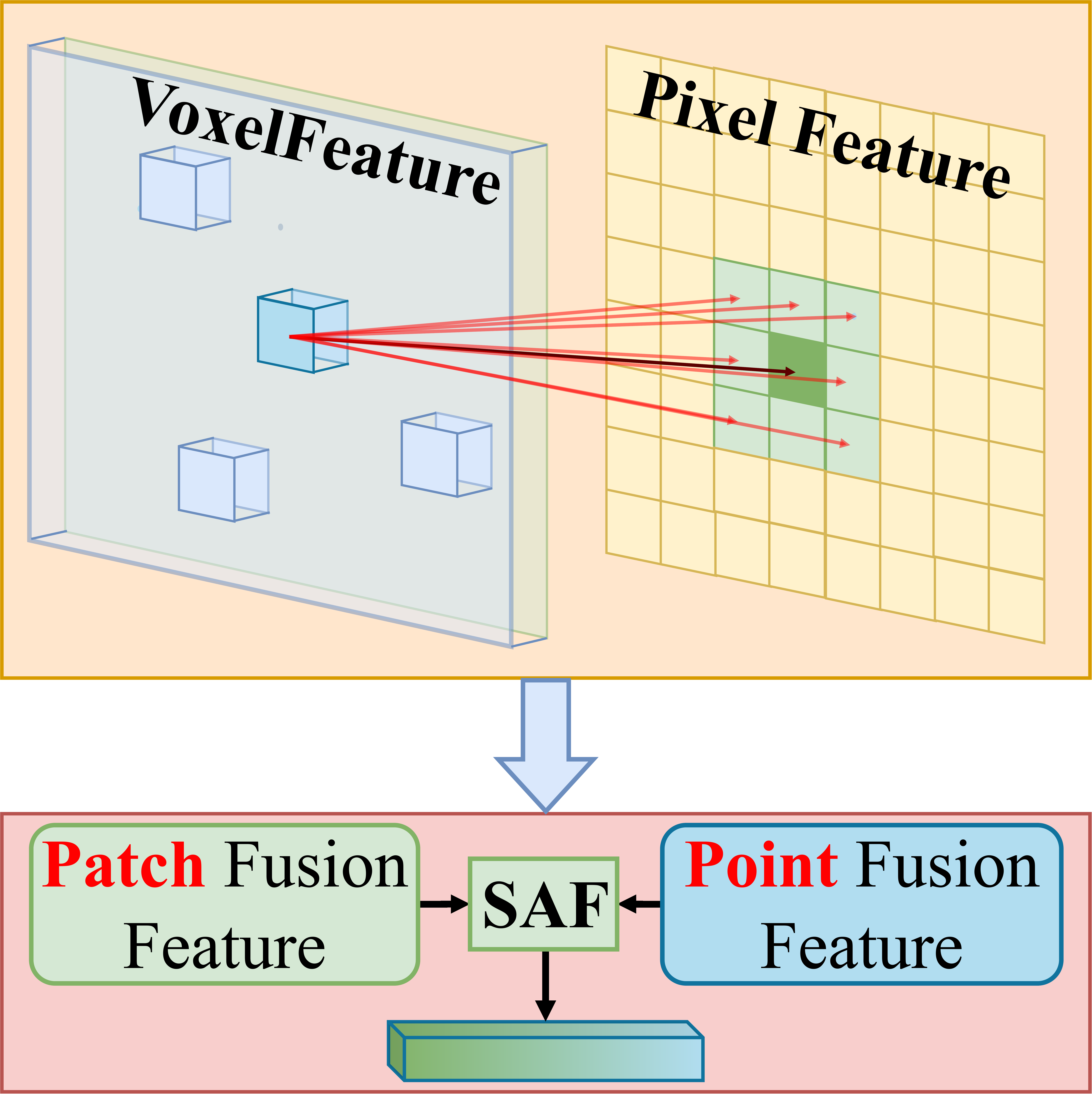}%
    \label{fig_second_case}}
    \subfloat[Performance]
    {\includegraphics[height=2.02in,width=1.91in]{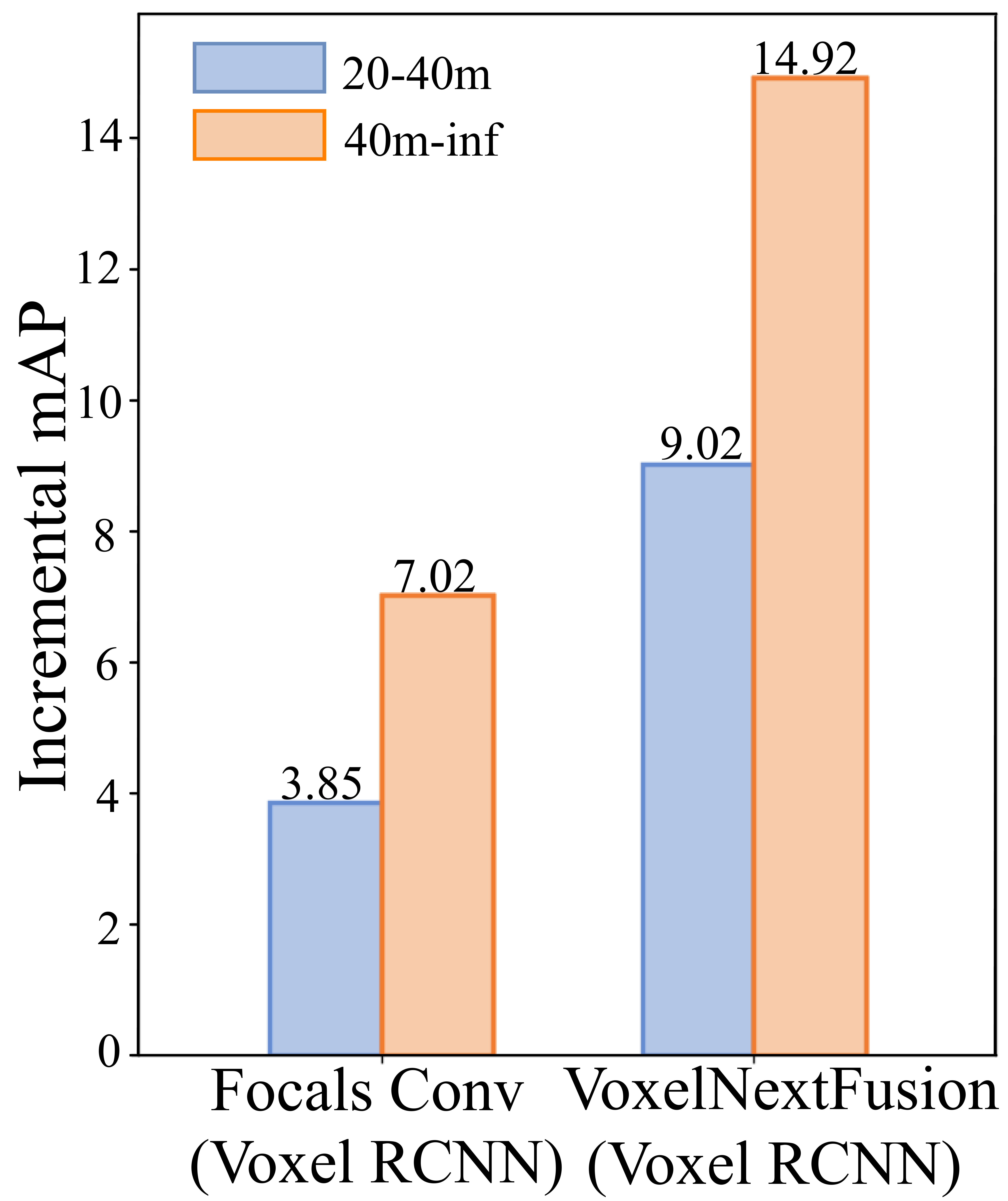}%
    \label{fig_third_case}}
    \caption{(a) To fuse point clouds and images accurately, state-of-the-art methods leverage one-to-one projection to correspond 3D-2D coordinates. However, due to the inconsistent resolution of the two modalities, for instance, in the case of a long-range object such as a car marked in 
    green, it contains 14 LiDAR points and more than 200 pixels. (b)To tackle this issue, we propose the \textcolor{black}{VoxelNextFusion} strategy that combines the one-to-many and one-to-one approaches to enlarge the usage of pixels. (c)The experiments demonstrate that our \textcolor{black}{VoxelNextFusion} significantly improves detection performance, particularly for long-range objects.}
    \label{fig_sim}
\end{figure*}

\begin{figure}[!t]
    \centering
    \includegraphics[width=\linewidth]{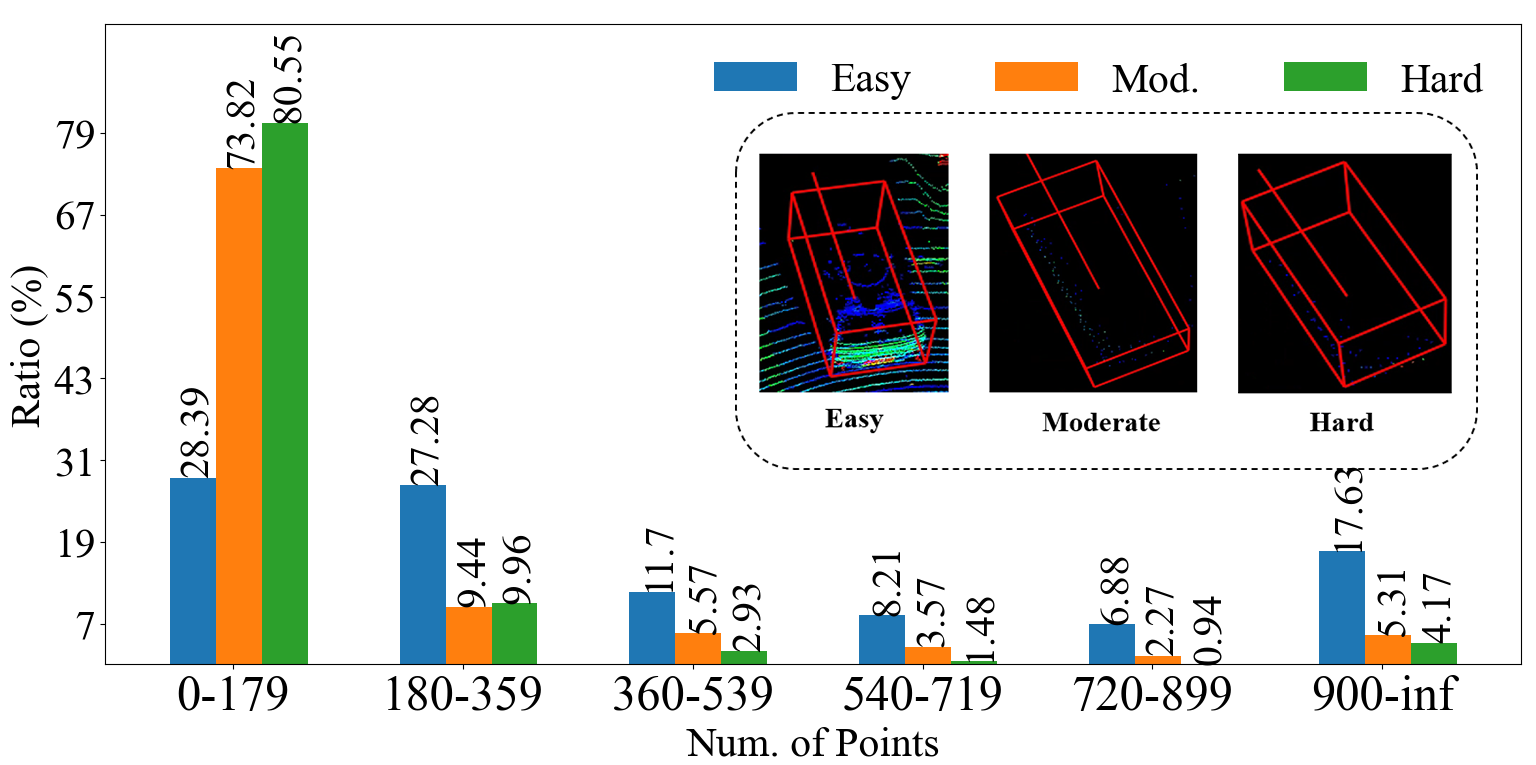}
    \caption{\textcolor{black}{Point Cloud Count Distribution by Difficulty Levels in KITTI GT Bounding Boxes. The data is sourced from the GT statistics of cars in the KITTI\cite{kitti} train dataset, comprising a total of 14,357 points. Among these, there are 3,153 points categorized as "easy," 4,893 points categorized as "moderate," and 2,971 points categorized as "hard." A lower point count within the GT bounding box indicates higher detection difficulty, with "Hard" cases being the most prevalent. As shown in Table \ref{tab_kitti_test}, a 3.20\% improvement on the "Hard" category demonstrates the effectiveness of our VoxelNextFusion.}}
    \label{fig:Kitti_easymodhard}
\end{figure}

\begin{abstract}
LiDAR-camera fusion can enhance the performance of 3D object detection by utilizing complementary information between depth-aware LiDAR points and semantically rich images. Existing voxel-based methods face significant challenges when fusing sparse voxel features with dense image features in a one-to-one manner, resulting in the loss of the advantages of images, including semantic and continuity information, leading to sub-optimal detection performance, especially at long distances. In this paper, we present \textcolor{black}{VoxelNextFusion}, a multi-modal 3D object detection framework specifically designed for voxel-based methods, which effectively bridges the gap between sparse point clouds and dense images. In particular, we propose a voxel-based image pipeline that involves projecting point clouds onto images to obtain both pixel- and patch-level features. These features are then fused using a self-attention to obtain a combined representation. Moreover, to address the issue of background features present in patches, we propose a feature importance module that effectively distinguishes between foreground and background features, thus minimizing the impact of the background features. Extensive experiments were conducted on the widely used KITTI and nuScenes 3D object detection benchmarks. Notably, our \textcolor{black}{VoxelNextFusion} achieved around +3.20\% in AP@0.7 improvement for car detection in hard level compared to the Voxel R-CNN baseline on the KITTI test dataset.
\end{abstract}

\begin{IEEEkeywords}
3D object detection, multi-modal fusion, patch fusion
\end{IEEEkeywords}

\IEEEpeerreviewmaketitle

\section{Introduction}

\IEEEPARstart
3D object detection is a critical task in autonomous driving and has been extensively studied with the develop of intelligent transportion system and 3D scene reconstruction technology~\cite{pursuing3D}. Although the availability of sensor data, such as cameras and LiDAR, has led to significant progress in single-modal 3D object detection, each modality has its shortcomings. LiDAR captures sparse point clouds that do not provide enough context to distinguish hard scenarios in distant or occluded areas. On the other hand, the camera produces RGB images that contain rich semantic information but lack depth information. Therefore, there are significant limitations in performance in single-modal scenarios. To overcome the limitations mentioned above, researchers have proposed multi-modal 3D object detection methods to leverage the complementary advantages between different modalities and improve detection performance.

Currently, most multi-modal 3D object detection methods~\cite{sfd, vff, focalconv, autoalign, deepfusion, vpfnet, epnet, frustum-pointnets} primarily rely on point cloud pipelines, with image pipelines serving as supplements. 
In this pattern, the point cloud branch uses point-based and voxel-based methods as the primary means of representation.
Voxel-based methods have been developed to adapt powerful RPN networks in 2D object detection. 
They transform irregular, unordered, and non-structured point cloud into structured data through voxelization processing, allowing for feature extraction by CNN. 
Although voxel-based multi-modal methods~\cite{sfd, vff, focalconv, autoalign, deepfusion, vpfnet} are very powerful, voxelization inevitably brings significant loss of information. The projection of voxel features onto image features utilizes a one-to-one mapping, as shown in Fig.~\ref{fig_first_case}, where each voxel feature is fused with a single pixel feature. 
However, this approach results in the loss of image semantics and continuity due to the fusion of a single voxel feature and a single pixel feature. 

The fundamental reason for the issues with voxel-based multi-modal methods~\cite{sfd, vff, focalconv, autoalign, deepfusion, vpfnet,graphalign,graphalign++} lies in the sparsity of point clouds, especially at long-range detection. In current mainstream outdoor 3D object detection datasets, such as KITTI\cite{kitti}, the projection of point clouds onto corresponding images reveals that only approximately 3\% of pixels have corresponding point cloud data. \textcolor{black}{The KITTI dataset categorizes the detection difficulty into three classes: "Easy," "Moderate," and "Hard." We have conducted a statistical analysis of the distribution of point cloud counts within Ground Truth (GT) bounding boxes for different difficulty levels in the KITTI dataset, as illustrated in Fig. \ref{fig:Kitti_easymodhard}. Notably, for "Moderate" and "Hard" objects, over 73\% and 80\%, respectively, have fewer than 180 points within their bounding boxes. Moreover, the "Hard" category encompasses smaller objects at long-range, characterized by highly incomplete shapes and structures, further challenging 3D object detection.}

One-to-one mapping is a fine-grained solution, which leads to an issue in voxel-based multi-modal methods~\cite{sfd, vff, focalconv, autoalign, deepfusion, vpfnet,graphalign,graphalign++}  that the voxel only uses one centroid for projection but a single voxel contains multiple points. Consequently, it makes the sparse point cloud even sparser after projecting. 
\textcolor{black}{As shown in Fig.~\ref{fig_first_case}, we illustrate the resolution mismatch caused by one-to-one mapping and observe that the pixel occupancy rate is only about 12\%. }
A \textcolor{black}{naive} strategy, if only to expand occupancy, is to map a voxel to multiple pixels via specific rules, which is a coarse-grained solution. 
Nevertheless, as our visualization results support, these mapped pixels are not equally important for the detection. 
Our findings indicate that the relevance of the mapped pixels for object detection is not uniformly distributed. This inconsistency arises due to the unequal correspondence between the information acquired from LiDAR depth features and each camera pixel. 
Specifically, certain pixels may incorporate irrelevant features such as environmental background elements, road surfaces, and vegetation, which do not contribute to the objective of object detection and hence can be deemed non-informative. 


To address the issues associated with existing voxel-based multi-modal 3D object detection, we propose a simple, unified, and effective multi-modal fusion framework, ~\textbf{\textcolor{black}{VoxelNextFusion}}. 
\textit{First}, we follow the principle of efficient fusion by proposing P$^{2}$-Fusion, which can fuse coarse- and fine-grained multimodal features while maintaining image continuity and high semantics. Without bells and whistles, it  uses a self attention for the fusion process.
\textit{Second}, we differentiate between foreground and background features to eliminate any potential impact of background pixel features, which further improves the exploitation of important features in the fusion process. 
\textit{Finally}, we conduct extensive experiments on two popular datasets KITTI~\cite{kitti} and nuScenes~\cite{nuscenes}. In the default setting, our method significantly improves the performance of most state-of-the-art methods. Our method demonstrates superior performance compared to previous methods on long-range objects, particularly when the target has a sparse point cloud, as evident in the KITTI Hard level benchmark in Table \ref{tab_kitti_test}, and the abation study of distance in Table\ref{tab_distances}.
\begin{figure*}[!t]
    \centering
    \includegraphics[width=1\linewidth]{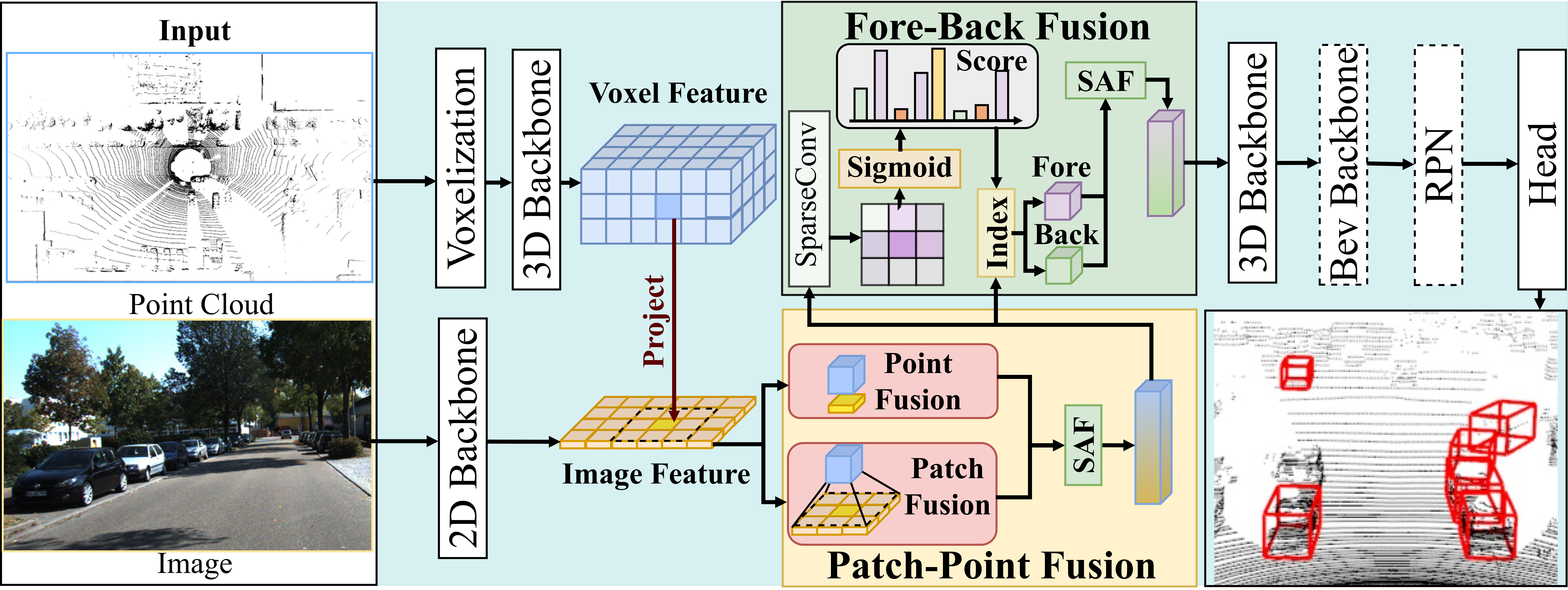}
    \caption{The framework of our \textcolor{black}{VoxelNextFusion}. First, we voxelized the points cloud and fed it into a 3D sparse convolution backbone. In the image branch, the image is fed into a 2D encoder. After that, we project the sparse voxel feature onto the image feature to conduct P$^2$-Fusion (Patch-Point Fusion) module. Second, we adopt the FB-Fusion (Foreground-Background Fusion) module that can weight features according to their foreground or background scores. Finally, the weighted feature is fed into a 3D convolution block and used to predict results. 'SAF' represents the self-attention Fusion module.}
    \label{fig:framework}
\end{figure*}

\section{Related work}
\subsection{3D Object Detection with Single Modality}

3D object detection is commonly conducted by using a single modality, either a camera or a LiDAR sensor \cite{wei2023surroundocc}. \textcolor{black}{Camera-based 3D detection methods~\cite{gao2022esgn, liu2020smoke, yang2023lite, yang2023mix, yang2023monogae,piccinelli2023idisc} take images as input and output object locations in a 3D manner. As early works, Deep3DBox~\cite{deep3d} and FCOS3D~\cite{fcos3d} transfer the 2D detection framework to 3D. 
SMOKE~\cite{liu2020smoke} proposes a concise framework via predicting keypoints to generate 3D objects. Recent works introduce geometric prior (\textit{e.g.,} 2D-3D keypoints~\cite{rtm3d}, adjacent object pairs~\cite{monopair}) to capture 3D cues.}
However, monocular cameras cannot provide accurate depth information. 
\textcolor{black}{Pseudo-LiDAR~\cite{pseudolidar}, as a pioneer in stereo-based method, leverages stereo camera construct image depth for generating pseudo-LiDAR points. BEVDepth~\cite{bevdepth} and BEVFormer~\cite{bevformer} utilize surrounding-view cameras to generate BEV feature with 3D cues.}

Although camera-based 3D object detection has made remarkable advances, its accuracy is far behind that of 3D object detection using LiDAR. 
In LiDAR-based detectors, point-based methods~\cite{pointrcnn, FARPXieTMM,3dssd} directly process irregular point clouds by PointNet series backbone~\cite{pointnet, pointnet++}. Voxel-based methods convert the point cloud into regular voxels~\cite{voxelnet,vpnet,casa,sievnet,hannet} or pillars~\cite{pointpillars}, which are convenient and high-efficiency for feature extraction using 3D or 2D CNN processing~\cite{second, voxset, pointpillars}.
Although LiDAR-based 3D object detection is superior to camera-based methods, it has limitations due to the sparse nature of point clouds, the lack of texture features, and poor semantic information \cite{wang2023multi,satgcn}. 
\subsection{3D Object Detection with Multi-modalities}

To address the limitations of single-modal, various methods combine the data from the two modalities to improve detection performance~\cite{wang2023multi}. 
PointPainting~\cite{pointpainting} strengthen 
LiDAR points with the semantic score of the corresponding camera pixel. 
PI-RCNN~\cite{pircnn} fuses semantic features from the image branch and RoI-wise LiDAR points in the refinement stage. 
Frustum PointNets~\cite{frustum-pointnets} and Frustum-ConvNet~\cite{frustum-convnet} utilize images to generate 2D proposals and then lift them up to 3D space (frustum) to narrow the searching space in point clouds. 
The MVX-Net~\cite{mvx-net} and EPNet~\cite{epnet} leverage one-to-one mapping strategy to index image features for LiDAR features and combine them. 
3D-CVF~\cite{3d-cvf} explores alignment strategies on feature maps across different modalities with a learned calibration matrix. \textcolor{black}{Some works leverage more in-depth fusion strategies, \textit{e.g.}, attention-based~\cite{deepinteraction,autoalignv2}, graph-based~\cite{graphrcnn}, to further improve cross-modal fusion performance. }
\textcolor{black}{Recent works~\cite{sfd,HMFI,bevfusion-mit,bevfusion-pku,MSMDFusion,virconv} lift 2D image to 3D representation for fuse LiDAR and camera in shared space and learn joint 3D representation. }
\textcolor{black}{Virtual point-based methods~\cite{sfd,virconv} introduce camera virtual points that can lead to dense multi-modal fusion.}
\textcolor{black}{However, the above methods leverage the calibration matrix to align the heterogeneous features, which have a risk of destroying the image semantic information and adjacency, thus restraining performance.} Other methods~\cite{deepfusion, autoalign, autoalignv2, transfusion} 
investigate a learnable alignment using the cross-attention mechanism. Although these methods effectively preserve the semantic information of images, the frequent query of image features by the attention mechanism increases computational costs. 

\section{\textcolor{black}{VoxelNextFusion}}
\label{sec:method}
In this section, we propose a simple, unified and effective multi-modal fusion framework that integrates coarse-grained and fine-grained point clouds and images to better facilitate voxel-based 3D object detection. Fig.~\ref{fig:framework} illustrates the architecture of our \textcolor{black}{VoxelNextFusion}. To achieve better fusion for voxel-based 3D object detection, we design two sub-modules, namely P$^2$-Fusion (Patch-Point Fusion) and FB-Fusion (Foreground-Background Fusion).

\subsection{Patch-Point Fusion}
Existing voxel-based multi-modal methods typically use a one-to-one mapping between voxels and images for fusion. While the camera pixel that uniquely corresponds to each voxel can be precisely located, LiDAR features represent a subset of points contained within a voxel, so their corresponding camera pixels lie within a polygon. The one-to-one mapping loses the original intention of using images, namely semantic and continuous properties, which is even worse for long-range detection. Therefore, we propose P$^2$-Fusion (Patch-Point Fusion) to compensate for the shortcomings. As shown in Fig.~\ref{fig:framework}, after voxelization of the original point cloud, multiple layers of 3D sparse convolution encoding are performed. We implement our proposed P$^2$-Fusion between the first and second layer encodings. P$^2$-Fusion is primarily composed of two stages: \textbf{Projection}, and \textbf{Fusion}.

\subsubsection{Projection}
\ 
In multi-modal 3D object detection, the core challenge is to align features for fusion. This is accomplished by utilizing a calibration matrix to transform the 3D coordinate system of voxels into the pixel coordinate system of images, thereby enabling the fusion of point clouds and image modalities. We project a 3D point cloud onto the image plane as follows:

\begin{equation}\label{equ3d2d}
z_{c}\left[\begin{array}{c}
u \\
v \\
1
\end{array}\right]=h \mathcal{K}\left[\begin{array}{ll}
R & T
\end{array}\right]\left[\begin{array}{c}
P_{x} \\
P_{y} \\
P_{z} \\
1
\end{array}\right]
\end{equation}
where, $P_{x}$, $P_{y}$, $P_{z}$ denote the LiDAR point's 3D locations, $u$, $v$ denote the corresponding 2D locations, and $z_{c}$ represents the depth of its projection on the image plane, $\mathcal{K}$ denotes the camera intrinsic parameter, $R$ and $T$ denote the rotation and the translation of the LiDAR with respect to the camera reference system, and $h$ denotes the scale factor due to down-sampling.

The raw image $\mathbf{I} \in \mathbb{R}^{W \times H \times 3 }$ is encoded by a pre-trained semantic segmenter DeepLabV3~\cite{DeepLabV3}, which generates an image feature $\mathbf{F_{I}} \in \mathbb{R}^{\frac{W}{4} \times \frac{H}{4} \times C_{I} }$, where $W$, $H$, and $C_{I}$ are the width of the image, the height of the image, and the channel number of the image feature, respectively. 
After the first layer of 3D sparse convolution, the  sparse encoding map is obtained, which consists of \textcolor{black}{the voxel feature $\mathbf{F_{v}} \in \mathbb{R}^{N \times C_{v} }$, voxel indices $\mathbf{V_{I}} \in \mathbb{R}^{N \times 3 }$}, where $N$, $C_{v}$, $3$ are the number of non-empty voxels, the channel number of the voxel feature, the coordinates of a point cloud represented by $(x,y,z)$ of the voxel space. The voxel indices $\mathbf{V_{I}}$ is transformed into 3D indexes $\mathbf{V_{3d}} \in \mathbb{R}^{N \times 3 }$ in the point cloud coordinate system as follows.

\begin{equation}\label{equ2}
\mathbf{S_{I}}=\mathbf{V_{I}}\times V_{stride}
\end{equation}
\begin{equation}\label{equ3}
\mathbf{V_{3d}}=\mathbf{S_{I}}\times V_{size} + R_{PC} 
\end{equation}
where, the addition operation is used, $C$, $C_{I}$, and $C_{v}$ are all equal. 

In the aforementioned context, $\mathbf{F_{I}} \in \mathbb{R}^{\frac{W}{4} \times \frac{H}{4} \times C_{I} }$ is obtained. Among them, where $C_{I}$ equals $C_{v}$, it is generally taken to be 16. However, $\frac{W}{4}$ and $\frac{H}{4}$ are not what we desire because the calibration file corresponds to $W$ and $H$ when projecting the point cloud onto the image. Thus, bilinear interpolation is employed for upsampling to achieve more accurately, resulting in the acquisition of a novel image feature $\mathbf{F^{'}_{I}} \in \mathbb{R}^{W \times H \times C_{I} }$. In order to mitigate the influence of feature misalignment resulting from point cloud data augmentation prior to 3D to 2D projection, a reverse transformation is applied to convert the point cloud back to its original coordinates, which involves operations such as undoing the flipping in the up-down direction. To complete that, we obtain the original point cloud coordinate system $\mathbf{V_{3d}} \in \mathbb{R}^{N \times 3 }$, and transform $\mathbf{V_{3d}}$ to pixel coordinates $\mathbf{V_{2d}} \in \mathbb{R}^{N \times  2}$ by the projection calibration matrix in
Equation (\ref{equ3d2d}), where 2 is the coordinates of the corresponding image pixels represented by $(x,y)$.

\begin{figure}[t]
    \centering
    \captionsetup[subfloat]{font=normalsize}
    \subfloat[One-to-One Fusion]
    {\includegraphics[width=1.40in,height=1.40in]{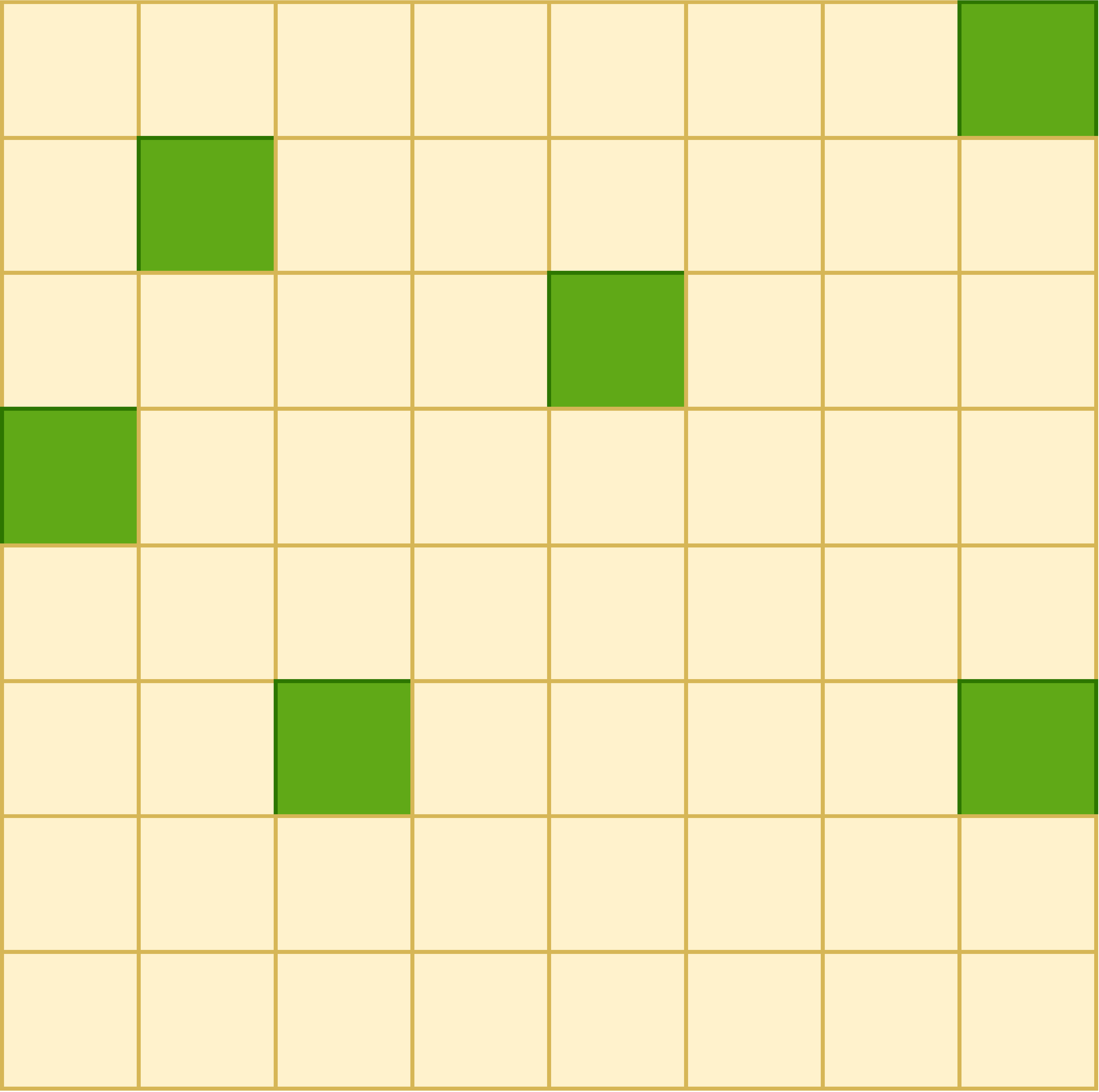}%
    \label{onetoonefusion}
    }
    \subfloat[One-to-Many Fusion]
    {\includegraphics[width=1.40in,height=1.40in]{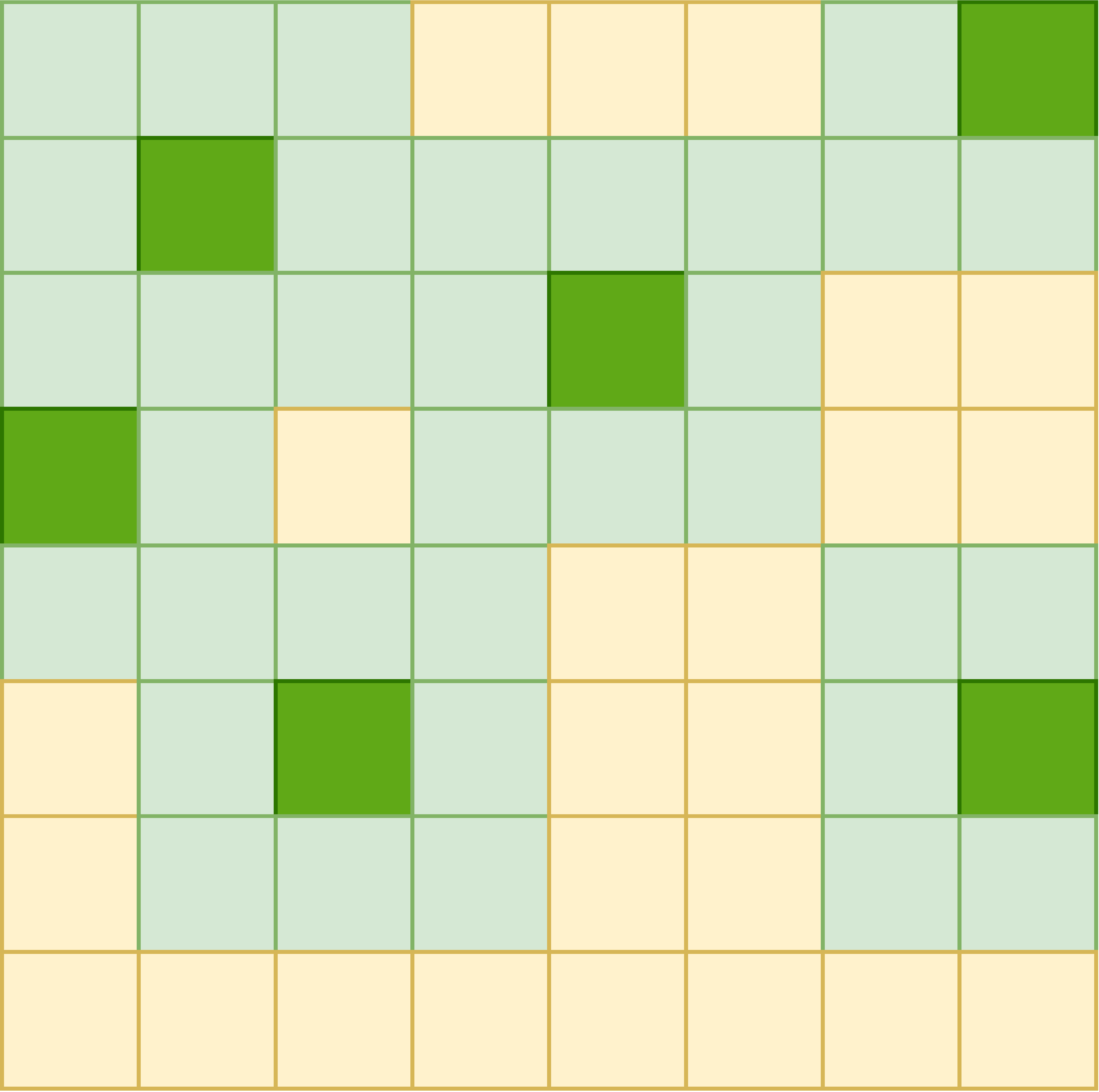}%
    \label{onetomanyfusion}}
    \caption{Comparison of one-to-one and one-to-many fusion. The 
    green square represents the features of the projected pixels, the
    yellow square represents the features of the unprojected pixels, and the 
    light green square represents the features of the neighboring pixels of the projected pixels.
    }
    \label{fusion_type}
\end{figure}
\begin{figure*}[!t]
    \centering
    \includegraphics[width=0.8\linewidth]{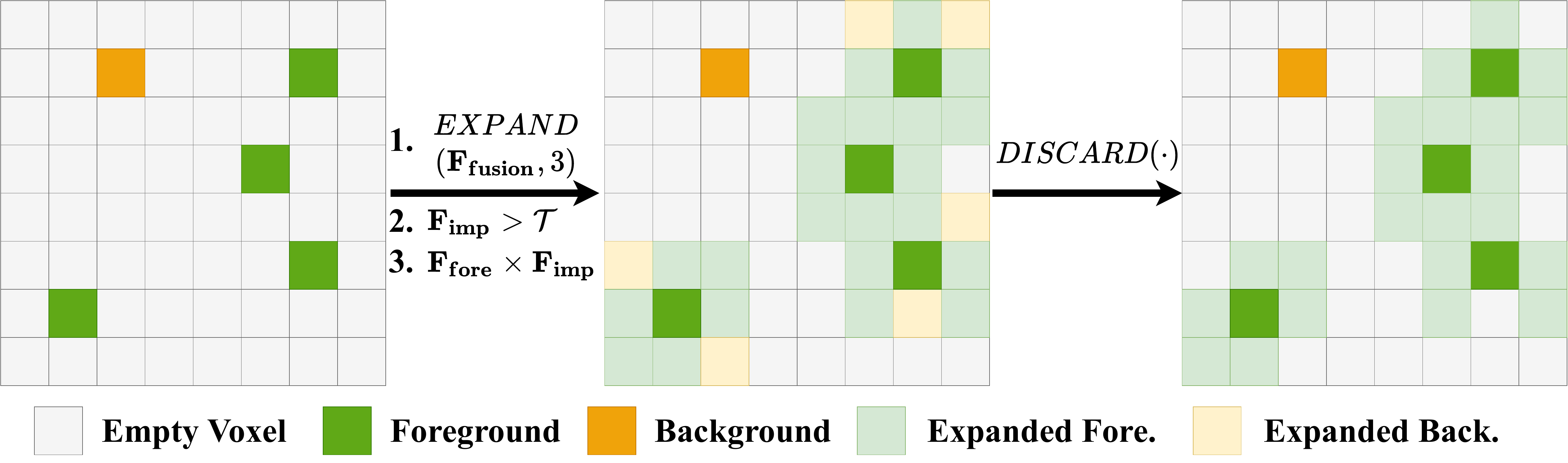}
    \caption{\textcolor{black}{Illustration of Splitting Foreground-Background. We note that this is a 2D example and can be easily extended to 3D cases.  Compared the $\mathbf{F_{imp}}$ with $\mathcal{T}$, we partition the 
    Foreground
    features and 
    Background
    features. To enhance the density of 
    Foreground
    features, we utilize the `EXPAND' operation to repeat the 
    Foreground
    features to their surroundings  ${K_{S}}^{3} - 1 $ neighbors. Compared the $\mathbf{F_{imp}}$ with $\mathcal{T}$, We discriminate between the 
    Expanded Fore.
    and 
    Expanded Back
    .  Subsequently, we employ the `DISCARD' operation to eliminate the 
    Expanded Back.}
    }
    \label{fig:fbfusion}
\end{figure*}

\subsubsection{Fusion}

\ 
One-to-one mapping accurately locates LiDAR points onto corresponding camera pixels, but it may suffer from data loss and imperfect geometric relations. One-to-many mapping can improve matching accuracy and accommodate sensor errors, but it increases computational complexity, requires careful weighting, and may be affected by occlusion. As shown in Fig.~\ref{fusion_type}, one-to-many mapping blends more pixel features, but not all pixels are effective and need to be filtered.
\ 
\newline 
\textbf{Point Fusion:}
In the aforementioned, we obtain the image feature $\mathbf{F^{'}_{I}} \in \mathbb{R}^{W \times H \times C_{I} }$ and the voxel feature  $\mathbf{F_{v}} \in \mathbb{R}^{N \times C_{v} }$, but their shapes are inconsistent, making fusion impossible. Therefore, by indexing the image feature $\mathbf{F^{'}_{I}} \in \mathbb{R}^{W \times H \times C_{I} }$ with pixel coordinates $\mathbf{V_{2d}} \in \mathbb{R}^{N \times  2}$, we obtain a novel image feature $\mathbf{F^{''}_{I}} \in \mathbb{R}^{N \times C_{I} }$ as follows. 
\begin{equation}
\label{equ4}
        \mathbf{F^{''}_{I}}=\{\mathbf{F^{'}_{I}}(\mathbf{V_{2d}}(i,0), \mathbf{V_{2d}}(i,1),:) \\| \forall i\in \left \{0,1,2\dots N-1 \right \} \}
\end{equation}


Now that the shapes of the voxel feature $\mathbf{F_{v}} \in \mathbb{R}^{N \times C_{v} }$ and the image feature $\mathbf{F^{''}_{I}} \in \mathbb{R}^{N \times C_{I} }$ are consistent, we can perform concatenation or addition operations to obtain the fused feature $\mathbf{F_{IV}} \in \mathbb{R}^{N \times C }$ as follows. 
\begin{equation}\label{equ5}
\mathbf{F_{IV}}=\mathbf{F^{''}_{I}}+ \mathbf{F_{v}}
\end{equation}
where, if the addition operation is used, $C$, $C_{I}$, and $C_{v}$ are all equal. If the concatenation operation is used, $C$ equals the concatenation of $C_{I}$ and $C_{v}$.
\
\newline 
\textbf{Patch Fusion:}
\begin{algorithm}[t]
\SetAlgoLined
\caption{Patch-Point Fusion workflow} 
\label{algorithm:PPFusion}

\KwIn{

\textcolor{black}{Image features $\mathbf{F^{'}_{I}} \in \mathbb{R}^{W \times H \times C_{I} }$.}

Voxel features $\mathbf{F_{v}} \in \mathbb{R}^{N \times C_{v} }$.


Point cloud range $R_{PC}$.

Hyper-parameters: No. of patch $K$.

}
\KwOut{

A noval fusion feature $\mathbf{F_{fusion}} \in \mathbb{R}^{N \times C }$.
}
\While{use image}{
 $\mathbf{V_{3d}} = TRANS(\mathbf{F_{v}}, R_{PC})$ 
 
 $\mathbf{V_{2d}}, \mathbf{V'_{2d}} = PROJECT(\mathbf{V_{3d}}, \mathbf{T}, K)$
 
 $\mathbf{F_{IV}} = PointFusion(\mathbf{F^{'}_{I}}, \mathbf{V_{2d}})$ 
 
 $\mathbf{F_{KIV}} = PatchFusion(\mathbf{F^{'}_{I}}, \mathbf{V'_{2d}})$ 
 
 $\mathbf{F_{fusion}} = SAF(\mathbf{F_{IV}}, \mathbf{F_{KIV}})$ 
}
\end{algorithm}

However, the one-to-one mapping approach used in Point Fusion results in sparsity of dense image feature. Therefore, a naive solution is to adopt a one-to-many mapping approach where a voxel feature is fused with neighboring pixels feature similar to the convolutional kernel. Specifically, our Patch Fusion solution for the one-to-many scenario is cleverly designed. As shown in Fig. \ref{onetomanyfusion}, it involves simply adding neighboring pixels to the pixel coordinates $\mathbf{V_{2d}} \in \mathbb{R}^{N \times  2}$, resulting in new pixel coordinates $\mathbf{V^{'}_{2d}} \in \mathbb{R}^{N \times K \times 2}$ that are enriched with neighbors as follows.
 
\begin{equation}\label{equ6}
\mathbf{V^{'}_{2d}} = \mathbf{V_{2d}}+ K_{off}
\end{equation}
\textcolor{black}{where, we utilize the broadcasting mechanism for both $\mathbf{V_{2d}}$ and $K_{off} \in \mathbb{R}^{K \times  2}$, where $K_{off}$ is the neighboring pixel coordinates.}

The subsequent procedure remains consistent with Point Fusion, whereby we retrieve image features $\mathbf{F_{KI}} \in \mathbb{R}^{N \times K \times C_{I} }$ with associated neighbors using a similar indexing approach as demonstrated in Equation (\ref{equ4}). Finally, as follow in Equation (\ref{equ5}), we obtain the fused feature $\mathbf{F_{KIV}} \in \mathbb{R}^{N \times K \times C }$ with pixel neighbors incorporated.

Finally, the workflow of fusion is demonstrated in Alg.~\ref{algorithm:PPFusion}. Generally, the incorporation of an image branch in multi-modal fusion methods increases their computational complexity compared to single-modal methods. 
\textcolor{black}{It is important to note that in Alg.~\ref{algorithm:PPFusion}, SAF refers to the SAF (Self-Attention Fusion) module. The SAF module includes operations such as MLP and self-attention
. 
For $\mathbf{F_{KIV}} \in \mathbb{R}^{N \times K \times C }$ and $\mathbf{F_{IV}} \in \mathbb{R}^{N  \times C }$, $\mathbf{F_{KIV}}$ is reshaped to $(N, K \times C)$, and $\mathbf{F_{IV}}$ undergoes a repeat operation, resulting in its shape being $(N, K \times C)$. Then, an addition operation is performed between $\mathbf{F_{KIV}}$ and $\mathbf{F_{IV}}$, and the result is passed through an MLP to obtain the fused feature $\mathbf{F_{C}} \in \mathbb{R}^{N  \times C }$. Subsequently, $\mathbf{F_{C}}$ enters a self-attention mechanism to obtain a new fused feature $\mathbf{F_{fusion}} \in \mathbb{R}^{N \times C }$.}

\begin{algorithm}[t]
\SetAlgoLined
\caption{Splitting Foreground-Background} \label{algorithm:FBFusion}
\KwIn{

The fused feature  $\mathbf{F_{fusion}}$.

Feature Importance $\mathbf{F_{imp}}$.


Importance threshold $\mathcal{T}$

}
\KwOut{

Foreground Features   $\mathbf{F_{Fore}}$

Background Features   $\mathbf{F_{Back}}$

}

\For{ $f$, $f_{imp}$ \textbf{in} $\mathbf{F_{fusion}}$, $\mathbf{F_{imp}}$}{

    $[f_{expand}^{imp},f_{fore}^{imp}] = f_{imp}$
    
    \eIf{$f_{fore}^{imp} > \mathcal{T}$}{ 
        $f_{fore} = f $
        
        
        \eIf{$f_{expand}^{imp} > \mathcal{T}$}{
            $f_{expand} = f_{expand}^{imp} \times f_{fore}$
            
            $f_{fore}^{dense} = Concat [f_{expand},  f_{fore}]$
        }{
            $DISCARD$;
        }
        }{
        $f_{back} = f$
        }
}

\end{algorithm}

\subsection{Foreground-Background Fusion}
The P$^2$-Fusion module combines one-to-many and one-to-one mappings where each voxel feature is represented by a patch feature containing multiple pixel features. Each LiDAR feature represents a subset of points in a voxel, and thus its corresponding camera pixel should be a polygon. Therefore, the one-to-many projection, i.e., patch fusion, is reasonable. However, it results in the problem of multiple pixels instead of a single pixel. One naive approach is to take the average of the pixel features in the patch. However, this is not a good strategy because the patch may contain background features such as roads, plants, or neighboring object features, hence, we need to identify key foreground features for detection while reducing the influence of background features and ensuring certain generalization. Therefore, we propose the FB-Fusion (Foreground-Background)  to further address the limitations of P$^2$-Fusion. Furthermore, our FB-Fusion module further densifies the foreground features to increase the density of sparse voxel features.

\textcolor{black}{ In the P$^2$-Fusion described above, we obtain the fused feature $\mathbf{F_{fusion}} \in \mathbb{R}^{N \times C }$. But $\mathbf{F_{fusion}}$ is too sparse and does not distinguish between foreground and background features. Therefore, we increase the denseness of foreground features by expanding their surrounding neighbors and distinguish foreground features from background features by evaluating the importance of voxel features, as shown in Fig. \ref{fig:fbfusion} and Alg.~\ref{algorithm:FBFusion}. To evaluate the importance of the voxel feature, we then employ the 3D submanifold convolution \cite{3DSemanticSegmentationWithSubmanifoldSparseConvNet, SubmanifoldSparseConvNet} and sigmoid function to process $\mathbf{F_{fusion}}$ for predicting the importance scores which includes itself and ${K_{S}}^{3} - 1 $ neighbours, denoted as $\mathbf{F_{imp}} \in \mathbb{R}^{N \times {K_{S}}^{3}} = Concat [\mathbf{F_{fore}^{imp}} \in \mathbb{R}^{N \times 1}, \mathbf{F_{expand}^{imp}} \in \mathbb{R}^{N \times {K_{S}}^{3}-1}]$ , where the kernel size is denote as $K_{S}$ and its common value is 3. 
If $\mathbf{F_{fore}^{imp}} > \mathcal{T}$, the corresponding features in $\mathbf{F_{fusion}}$ are considered as foreground features $\mathbf{F_{fore}} \in \mathbb{R}^{\alpha \times C}$; otherwise, they are regarded as background features $\mathbf{F_{back}} \in \mathbb{R}^{\beta  \times C}$. Where $\alpha + \beta = N$, where $\alpha$ represents the number of foreground voxels, and $\beta$ represents the number of background voxels.
As depicted in Fig. \ref{fig:fbfusion}, we employ the EXPAND operation to replicate voxel features $\mathbf{F_{fore}}$ onto ${K_{S}}^3-1$ neighboring voxels. Subsequently, the corresponding $\mathbf{F_{expand}^{imp}}$ values for these neighboring voxels are compared to a threshold $\mathcal{T}$. If $\mathbf{F_{expand}^{imp}} < \mathcal{T}$, they are classified as Expanded Background; if $\mathbf{F_{expand}^{imp}} > \mathcal{T}$, they are regarded as Expanded Foreground. The features of the Expanded Foreground are represented as $\mathbf{F}_{expand} \in \mathbb{R}^{N, {K_{S}}^3-1, C} = \mathbf{F_{fore}}  \times \mathbf{F_{expand}^{imp}}$. In this case, we use the DISCARD operation to discard the Expanded Background by treating it as empty voxels. In addition, we combine $\mathbf{F}_{expand}$ and $\mathbf{F}_{fore}$ to obtain dense foreground features $\mathbf{F}_{fore}^{dense}$ as follows.
\begin{equation}
    \mathbf{F}_{fore}^{dense} = Concat[\mathbf{F}_{expand}, \mathbf{F}_{fore}]
\end{equation}
Finally, we separated out foreground $\mathbf{F}_{fore}^{dense}$ and background features $\mathbf{F}_{back}$, and expanded and weighted the importance of foreground features, meaning that informative foreground features are enhanced. Then, we feed $\mathbf{F}_{fore}^{dense}$ and $\mathbf{F}_{back}$ into the SAF module to obtain the newly fused feature, which is subsequently incorporated into the 3D Backbone.}


\section{Experiments}\label{Experiments}

In this section, we present the details of each dataset and the experimental setup of \textcolor{black}{VoxelNextFusion}, and evaluate the performance of 3D object detection on KITTI~\cite{kitti} and nuScenes~\cite{nuscenes} datasets.

\subsection{Dataset and Evaluation Metrics}\label{sectionIV-A}
\subsubsection{KITTI dataset} 
The KITTI dataset~\cite{kitti} provides synchronized LiDAR point clouds and front-view camera images. It consists of 7,481 training samples and 7,518 test samples. \textcolor{black}{As a common practice \cite{voxelrcnn,focalconv,pvrcnn}, the training data are divided into a train set with 3712 samples and a val set with 3769 samples to conduct evaluation on the \textit{val} set. To perform evaluation on the \textit{test} dataset using the official KITTI test server, we follow the approach outlined in PV-RCNN~\cite{pvrcnn}. Our model is trained with 80\% of the 7,481 training samples, which amounts to 5,985 samples.}
The standard evaluation metric for object detection is the mean Average Precision (mAP), computed using recall at 40 positions (R40). In this work, we evaluate our models on the most commonly used the \textcolor{black}{Car, Pedestrian, and Cyclist} using Average Precision (AP) with an Intersection over Union (IoU) threshold of  \textcolor{black}{0.7, 0.5, and 0.5}, respectively. 
\subsubsection{nuScenes dataset} 
The nuScenes dataset~\cite{nuscenes} is a large-scale 3D detection benchmark consisting of 700 training scenes, 150 validation scenes, and 150 testing scenes. The data were collected using six multi-view cameras and a 32-beam LiDAR sensor. 
It includes 360-degree object annotations for 10 object classes. To evaluate the detection performance, the primary metrics used 
are the mean Average Precision (mAP) and \textcolor{black}{the nuScenes detection score (NDS)~\cite{nuscenes}, which assess detection accuracy in terms of classification, bounding box location, size, orientation, attributes, and velocity.}
\textcolor{black}{For efficiently conducting the ablation experiments, we randomly divided the 700 training scenes into subsets of 70 (representing $\frac{1}{10}$ of the data) and 175 (representing $\frac{1}{4}$ of the data) and all results are evaluated on the full validation set.}

\begin{table*}[t]
\scriptsize
\caption{Performance comparison with the SOTA methods on KITTI \textcolor{red}{\textit{test}} set. The (Car, Pedestrian, Cyclist) results are reported by the AP with (0.7,0.5,0.5) IoU threshold and 40 recall points. ‘L’ and ‘C’ represent LiDAR and Camera, respectively.}
\renewcommand\arraystretch{1}
\setlength{\tabcolsep}{0.5mm}{

\begin{tabular*}{\linewidth}{l|c|c|c|c|c|c|c|c|c|c|c|c|c|c|c|c|c|c|c}
\toprule
               &          & \multicolumn{6}{c|}{Car}                                                                                                                       & \multicolumn{6}{c|}{Pedestrian}                                                                                                                & \multicolumn{6}{c}{Cyclist}                                                                                                                    \\ \cmidrule(lr){3-20} 
Method         & Modality & \multicolumn{3}{c|}{$AP_{3D}$ ($\%$)}                                                        & \multicolumn{3}{c|}{$AP_{BEV}$ ($\%$)}                                  & \multicolumn{3}{c|}{$AP_{3D}$ ($\%$)}                                                        & \multicolumn{3}{c|}{$AP_{BEV}$ ($\%$)}                                  & \multicolumn{3}{c|}{$AP_{3D}$ ($\%$)}                                                        & \multicolumn{3}{c}{$AP_{BEV}$ ($\%$)}                                   \\ \cmidrule(lr){3-20} 
               &          & \multicolumn{1}{l|}{Easy} & \multicolumn{1}{l|}{Mod} & \multicolumn{1}{c|}{Hard} & \multicolumn{1}{c|}{Easy} & \multicolumn{1}{c|}{Mod} & Hard & \multicolumn{1}{c|}{Easy} & \multicolumn{1}{c|}{Mod} & \multicolumn{1}{c|}{Hard} & \multicolumn{1}{c|}{Easy} & \multicolumn{1}{c|}{Mod} & Hard & \multicolumn{1}{c|}{Easy} & \multicolumn{1}{c|}{Mod} & \multicolumn{1}{c|}{Hard} & \multicolumn{1}{c|}{Easy} & \multicolumn{1}{c|}{Mod} & Hard \\ \midrule

BSAODet \cite{xiao2023balanced}                & L                         & \multicolumn{1}{c|}{88.89}          & \multicolumn{1}{c|}{81.74}          & 77.14           & \multicolumn{1}{c|}{-}          & \multicolumn{1}{c|}{-}          & -      & 51.71    & 43.63   & 41.09   & -    & -   & -    & 82.65    & \textbf{67.79}   & \textbf{60.26}    & -    & -   & -    \\
H$^2$3D R-CNN \cite{deng2021multi}                & L                         & \multicolumn{1}{c|}{90.43}          & \multicolumn{1}{c|}{81.55}          & 77.22           & \multicolumn{1}{c|}{92.85}          & \multicolumn{1}{c|}{88.87}          & 86.07    & 52.75    & 45.26   & 41.56    & 58.14    & 50.43   & 46.72    & 78.67    & 62.74   & 55.78    & 82.76    & 67.90   & 60.49      \\
SIEV-Net\cite{sievnet} & L & \multicolumn{1}{c|}{85.21} & \multicolumn{1}{c|}{76.18} & \multicolumn{1}{c|}{70.60} & - & - & - & 54.00 & 44.80 & 41.11 & - & - & - & 78.75 & 59.99 & 52.37 & - & - & - \\
PointPillars \cite{pointpillars}           & L                         & \multicolumn{1}{c|}{82.58}          & \multicolumn{1}{c|}{74.31}          & 68.99          & \multicolumn{1}{c|}{90.07}          & \multicolumn{1}{c|}{86.56}          & 82.81   & \multicolumn{1}{c|}{51.45}    & \multicolumn{1}{c|}{41.92}   & \multicolumn{1}{c|}{38.89}    & \multicolumn{1}{c|}{57.60}    & \multicolumn{1}{c|}{48.64}   & 45.78    & \multicolumn{1}{c|}{77.10}    & \multicolumn{1}{c|}{58.65}   & \multicolumn{1}{c|}{51.92}    & \multicolumn{1}{c|}{79.90}    & \multicolumn{1}{c|}{62.73}   & 55.58       \\

VoxSet \cite{voxset}                 & L                         & \multicolumn{1}{c|}{88.53}          & \multicolumn{1}{c|}{82.06}          & 77.46          & \multicolumn{1}{c|}{-}              & \multicolumn{1}{c|}{-}              & -      & \multicolumn{1}{c|}{-}    & \multicolumn{1}{c|}{-}   & \multicolumn{1}{c|}{-}    & \multicolumn{1}{c|}{-}    & \multicolumn{1}{c|}{-}   & -    & \multicolumn{1}{c|}{-}    & \multicolumn{1}{c|}{-}   & \multicolumn{1}{c|}{-}    & \multicolumn{1}{c|}{-}    & \multicolumn{1}{c|}{-}   & -        \\
TANet \cite{tanet}                  & L                         & \multicolumn{1}{c|}{83.81}          & \multicolumn{1}{c|}{75.38}          & 67.66          & \multicolumn{1}{c|}{-}          & \multicolumn{1}{c|}{-}          & -    & \multicolumn{1}{c|}{\textbf{54.92}}    & \multicolumn{1}{c|}{46.67}   & \multicolumn{1}{c|}{\textbf{42.42} }    & \multicolumn{1}{c|}{-}    & \multicolumn{1}{c|}{-}   & -    & \multicolumn{1}{c|}{73.84}    & \multicolumn{1}{c|}{59.86}   & \multicolumn{1}{c|}{53.46}    & \multicolumn{1}{c|}{-}    & \multicolumn{1}{c|}{-}   & -       \\
MMF \cite{mmf}                    & L\&C                      & \multicolumn{1}{c|}{86.81}          & \multicolumn{1}{c|}{76.75}          & 68.41          & \multicolumn{1}{c|}{89.49}          & \multicolumn{1}{c|}{87.47}          & 79.10     & \multicolumn{1}{c|}{-}    & \multicolumn{1}{c|}{-}   & \multicolumn{1}{c|}{-}    & \multicolumn{1}{c|}{-}    & \multicolumn{1}{c|}{-}   & -    & \multicolumn{1}{c|}{-}    & \multicolumn{1}{c|}{-}   & \multicolumn{1}{c|}{-}    & \multicolumn{1}{c|}{-}    & \multicolumn{1}{c|}{-}   & -      \\

PI-RCNN \cite{pircnn}                & L\&C                      & \multicolumn{1}{c|}{84.37}          & \multicolumn{1}{c|}{74.82}          & 70.03          & \multicolumn{1}{c|}{-}          & \multicolumn{1}{c|}{-}          & -     & \multicolumn{1}{c|}{-}    & \multicolumn{1}{c|}{-}   & \multicolumn{1}{c|}{-}    & \multicolumn{1}{c|}{-}    & \multicolumn{1}{c|}{-}   & -    & \multicolumn{1}{c|}{-}    & \multicolumn{1}{c|}{-}   & \multicolumn{1}{c|}{-}    & \multicolumn{1}{c|}{-}    & \multicolumn{1}{c|}{-}   & -      \\
EPNet \cite{epnet}                  & L\&C                      & \multicolumn{1}{c|}{89.81}          & \multicolumn{1}{c|}{79.28}          & 74.59          & \multicolumn{1}{c|} {94.22}          & \multicolumn{1}{c|}{88.47}          & 83.69    & \multicolumn{1}{c|}{-}    & \multicolumn{1}{c|}{-}   & \multicolumn{1}{c|}{-}    & \multicolumn{1}{c|}{-}    & \multicolumn{1}{c|}{-}   & -    & \multicolumn{1}{c|}{-}    & \multicolumn{1}{c|}{-}   & \multicolumn{1}{c|}{-}    & \multicolumn{1}{c|}{-}    & \multicolumn{1}{c|}{-}   & -        \\
PointPainting \cite{pointpainting}          & L\&C                      & \multicolumn{1}{c|}{82.11}          & \multicolumn{1}{c|}{71.70}          & 67.08          & \multicolumn{1}{c|}{-}              & \multicolumn{1}{c|}{-}              & -       & \multicolumn{1}{c|}{50.32}    & \multicolumn{1}{c|}{40.97}   & \multicolumn{1}{c|}{37.84}    & \multicolumn{1}{c|}{-}    & \multicolumn{1}{c|}{-}   & -    & \multicolumn{1}{c|}{77.63}    & \multicolumn{1}{c|}{63.78}   & \multicolumn{1}{c|}{55.89}    & \multicolumn{1}{c|}{-}    & \multicolumn{1}{c|}{-}   & -         \\
Fast-CLOCs \cite{fastclocs}             & L\&C                      & \multicolumn{1}{c|}{89.11}          & \multicolumn{1}{c|}{80.34}          & 76.98          & \multicolumn{1}{c|}{93.02}          & \multicolumn{1}{c|}{89.49}          & 86.39      & \multicolumn{1}{c|}{52.10}    & \multicolumn{1}{c|}{42.72}   & \multicolumn{1}{c|}{39.08}    & \multicolumn{1}{c|}{57.19}    & \multicolumn{1}{c|}{48.27}   & 44.55    & \multicolumn{1}{c|}{\textbf{82.83}}    & \multicolumn{1}{c|}{65.31}   & \multicolumn{1}{c|}{57.43}    & \multicolumn{1}{c|}{83.34}    & \multicolumn{1}{c|}{67.55}   & 59.61    \\
Focals Conv-F \cite{focalconv}            & L\&C                      & \multicolumn{1}{c|}{90.55}          & \multicolumn{1}{c|}{82.28}          & 77.59          & \multicolumn{1}{c|}{-}              & \multicolumn{1}{c|}{-}              & -      & \multicolumn{1}{c|}{-}    & \multicolumn{1}{c|}{-}   & \multicolumn{1}{c|}{-}    & \multicolumn{1}{c|}{-}    & \multicolumn{1}{c|}{-}   & -    & \multicolumn{1}{c|}{-}    & \multicolumn{1}{c|}{-}   & \multicolumn{1}{c|}{-}    & \multicolumn{1}{c|}{-}    & \multicolumn{1}{c|}{-}   & -          \\
Graph-VoI\cite{graphr-cnn}            & L\&C                      & \multicolumn{1}{c|}{\textbf{91.89}}          & \multicolumn{1}{c|}{83.27}          & 77.78          & \multicolumn{1}{c|}{\textbf{95.69}}              & \multicolumn{1}{c|}{90.10}              & 86.85      & \multicolumn{1}{c|}{-}    & \multicolumn{1}{c|}{-}   & \multicolumn{1}{c|}{-}    & \multicolumn{1}{c|}{-}    & \multicolumn{1}{c|}{-}   & -    & \multicolumn{1}{c|}{-}    & \multicolumn{1}{c|}{-}   & \multicolumn{1}{c|}{-}    & \multicolumn{1}{c|}{-}    & \multicolumn{1}{c|}{-}   & -          \\
SFD\cite{sfd}            & L\&C                      & \multicolumn{1}{c|}{91.73}          & \multicolumn{1}{c|}{\textbf{84.76}}          & 77.92          & \multicolumn{1}{c|}{95.64}              & \multicolumn{1}{c|}{\textbf{91.85}}              & 86.83      & \multicolumn{1}{c|}{-}    & \multicolumn{1}{c|}{-}   & \multicolumn{1}{c|}{-}    & \multicolumn{1}{c|}{-}    & \multicolumn{1}{c|}{-}   & -    & \multicolumn{1}{c|}{-}    & \multicolumn{1}{c|}{-}   & \multicolumn{1}{c|}{-}    & \multicolumn{1}{c|}{-}    & \multicolumn{1}{c|}{-}   & -          \\
EPNet++\cite{epnet++}            & L\&C                      & \multicolumn{1}{c|}{91.37}          & \multicolumn{1}{c|}{81.96}          & 76.71          & \multicolumn{1}{c|}{-}              & \multicolumn{1}{c|}{-}              & -      & \multicolumn{1}{c|}{52.79}    & \multicolumn{1}{c|}{44.38}   & \multicolumn{1}{c|}{41.29}    & \multicolumn{1}{c|}{-}    & \multicolumn{1}{c|}{-}   & -    & \multicolumn{1}{c|}{76.15}    & \multicolumn{1}{c|}{59.71}   & \multicolumn{1}{c|}{53.67}    & \multicolumn{1}{c|}{-}    & \multicolumn{1}{c|}{-}   & -          \\
\midrule
Voxel R-CNN \cite{voxelrcnn}           & L                         & \multicolumn{1}{c|}{90.90}          & \multicolumn{1}{c|}{81.62}          & 77.06          & \multicolumn{1}{c|}{-}              & \multicolumn{1}{c|}{-}              & -       & \multicolumn{1}{c|}{-}    & \multicolumn{1}{c|}{-}   & \multicolumn{1}{c|}{-}    & \multicolumn{1}{c|}{-}    & \multicolumn{1}{c|}{-}   & -    & \multicolumn{1}{c|}{-}    & \multicolumn{1}{c|}{-}   & \multicolumn{1}{c|}{-}    & \multicolumn{1}{c|}{-}    & \multicolumn{1}{c|}{-}   & -         \\
Voxel R-CNN*             & L                         & \multicolumn{1}{c|}{90.76}          & \multicolumn{1}{c|}{81.69}          & 77.42          & \multicolumn{1}{c|}{92.89}          & \multicolumn{1}{c|}{89.97}          & 84.69      & \multicolumn{1}{c|}{52.57}    & \multicolumn{1}{c|}{44.86}   & \multicolumn{1}{c|}{39.09}    & \multicolumn{1}{c|}{57.66}    & \multicolumn{1}{c|}{49.32}   & 44.15    & \multicolumn{1}{c|}{77.54}    & \multicolumn{1}{c|}{64.00}   & \multicolumn{1}{c|}{53.15}    & \multicolumn{1}{c|}{79.68}    & \multicolumn{1}{c|}{67.56}   & 62.70      \\
\rowcolor{blue!10} + \textbf{\textcolor{black}{VoxelNextFusion}}          & L\&C                      & \multicolumn{1}{c|}{90.90} & \multicolumn{1}{c|}{82.93} & \textbf{80.62}\textit{\fontsize{6}{0}\selectfont\textcolor{red}{+3.20}} & \multicolumn{1}{c|}{94.46}          & \multicolumn{1}{c|}{90.73} & 88.34\textit{\fontsize{6}{0}\selectfont\textcolor{red}{+3.65}}   & \multicolumn{1}{c|}{53.27}    & \multicolumn{1}{c|}{\textbf{47.86}}   & \multicolumn{1}{c|}{42.11\textit{\fontsize{6}{0}\selectfont\textcolor{red}{+3.02}}}    & \multicolumn{1}{c|}{57.82}    & \multicolumn{1}{c|}{\textbf{51.48}}   & 45.89\textit{\fontsize{6}{0}\selectfont\textcolor{red}{+1.74}}    & \multicolumn{1}{c|}{78.56}    & \multicolumn{1}{c|}{65.27}   & \multicolumn{1}{c|}{54.24\textit{\fontsize{6}{0}\selectfont\textcolor{red}{+1.09}}}    & \multicolumn{1}{c|}{80.00}    & \multicolumn{1}{c|}{68.81}   & 63.51\textit{\fontsize{6}{0}\selectfont\textcolor{red}{+0.81}}  \\ 
\midrule
PV-RCNN \cite{pvrcnn}                & L                         & \multicolumn{1}{c|}{90.25}          & \multicolumn{1}{c|}{81.43}          & 76.82          & \multicolumn{1}{c|}{94.98}          & \multicolumn{1}{c|}{90.65}          & 86.14    & 52.17     & 43.29      & 40.29   & 59.86     & 50.57     & 46.74     & 78.60     & 63.71     & 57.65     & 82.49         & 68.89     & 62.41    \\
PV-RCNN *                & L                         & \multicolumn{1}{c|}{90.61}          & \multicolumn{1}{c|}{81.51}          & 76.81          & \multicolumn{1}{c|}{94.68}          & \multicolumn{1}{c|}{90.87}          & 86.19    & 52.10    & 43.63      & 40.44   & 60.06     & 50.43     & 46.81     & 78.58     & 63.83     & 57.71     & 82.50         & 68.93     & 62.57    \\
\rowcolor{blue!10} + \textbf{\textcolor{black}{VoxelNextFusion}}          & L\&C                      & \multicolumn{1}{c|}{90.40} & \multicolumn{1}{c|}{82.03} & 79.86\textit{\fontsize{6}{0}\selectfont\textcolor{red}{+3.05}} & \multicolumn{1}{c|}{94.97}          & \multicolumn{1}{c|}{91.31} & \textbf{89.06\textit{\fontsize{6}{0}\selectfont\textcolor{red}{+2.87}}}   & \multicolumn{1}{c|}{52.56}    & \multicolumn{1}{c|}{45.72}   & \multicolumn{1}{c|}{41.85\textit{\fontsize{6}{0}\selectfont\textcolor{red}{+1.41}}}    & \multicolumn{1}{c|}{\textbf{61.71} }    & \multicolumn{1}{c|}{51.30}   & \textbf{47.89\textit{\fontsize{6}{0}\selectfont\textcolor{red}{+1.07}}}     & \multicolumn{1}{c|}{79.28}    & \multicolumn{1}{c|}{64.47}   & \multicolumn{1}{c|}{58.25\textit{\fontsize{6}{0}\selectfont\textcolor{red}{+0.54}}}    & \multicolumn{1}{c|}{83.00}    & \multicolumn{1}{c|}{\textbf{69.93} }   & \textbf{63.71\textit{\fontsize{6}{0}\selectfont\textcolor{red}{+1.14}}}    \\ 
\bottomrule
\end{tabular*}
\label{tab_kitti_test}
\begin{tablenotes}
\footnotesize
\item[1] * denotes re-implement result.
\item[1] The color \textcolor{red}{red} indicates improvement.
\end{tablenotes}
}
\end{table*}
\begin{table}[t]
\scriptsize
\centering
\caption{Performance comparison with the SOTA methods on KITTI \textit{val} set for \textcolor{black}{car category}. The results are reported by the AP with 0.7 IoU threshold and 40 recall points. ‘L’ and ‘C’ represent LiDAR and Camera, respectively.}
\renewcommand\arraystretch{1}
\resizebox{\linewidth}{!}{
\begin{tabular}{l|c|c|c|c|c|c|c}
\toprule
\multirow{3}{*}{Method} & \multirow{3}{*}{Moiality} & \multicolumn{3}{c|}{$AP_{3D}$ (\%)}   & \multicolumn{3}{c}{$AP_{BEV}$ (\%)}                                                                                                                      \\ \cmidrule(lr){3-8} 
                        &                           & Easy           & Mod.           & Hard    & Easy           & Mod.           & Hard            \\ \midrule
PointRCNN \cite{pointrcnn}              & L                         & 88.88          & 78.63          & 77.38    & -             & -             & -           \\
H$^2$3D R-CNN \cite{deng2021multi}              & L                         & 89.63          & 85.20          & 79.08    & -             & -             & -           \\
MedTr-TSD \cite{tian2023medoidsformer}                 & L                         &  89.27        &  84.24          & 78.85   & -             & -             & -                \\
CT3D \cite{ct3d}                   & L                         & \textbf{92.85} & 85.82          & 83.46    & 96.14	& 91.88	 & 89.63
          \\

Voxel R-CNN \cite{voxelrcnn}             & L                         & 92.38          & 85.29          & 82.86   & 95.52          & 91.25          & 88.99            \\
PV-RCNN \cite{pvrcnn}                & L                         & 92.57          & 84.83          & 82.69   & 95.76 & 91.11          & 88.93              \\
CasA \cite{casa}                & L                         & 92.73          & 85.89          & 83.57   & - & -          & -              \\
MV3D \cite{MV3D}                   & L\&C                      & 71.29          & 62.68          & 56.56    & 86.55 	& 78.10 	& 76.67 
           \\

MMF \cite{mmf}                    & L\&C                      & 87.90          & 77.87          & 75.57       & \textbf{96.66}             & 88.25             & 79.60         \\
PI-RCNN \cite{pircnn}                & L\&C                      & 88.27          & 78.53          & 77.75     & -             & -             & -             \\
EPNet \cite{epnet}                  & L\&C                      & 92.28          & 82.59          & 80.14     & 95.51 	& 88.76 	& 88.36 \\
\midrule        
Voxel R-CNN $^*$            & L                         & 92.32          & 85.06          & 82.80   & 95.48          & 91.06          & 89.06            \\
\rowcolor{blue!10} +our \textbf{\textcolor{black}{VoxelNextFusion}}          & L\&C                      & 92.78          & \textbf{86.89} & 84.59\textit{\fontsize{6}{0}\selectfont\textcolor{red}{+1.79}}  & 95.74  & \textbf{92.87}   & \textbf{91.09}\textit{\fontsize{6}{0}\selectfont\textcolor{red}{+2.03}}  \\ \midrule
PV-RCNN $^*$                & L                         & 92.53          & 84.80          & 82.71   & 95.70 & 91.19          & 89.00              \\
\rowcolor{blue!10} +our \textbf{\textcolor{black}{VoxelNextFusion}}                  & L\&C                         & 92.43         & 85.61         & \textbf{84.70}\textit{\fontsize{6}{0}\selectfont\textcolor{red}{+1.99}}   & 95.54 & 91.25         & 90.93\textit{\fontsize{6}{0}\selectfont\textcolor{red}{+1.93}}              \\ 
\bottomrule
\end{tabular}}
\label{tab_kitti_val}
\begin{tablenotes}
\footnotesize
\item[1] * denotes re-implement result.
\item[1] The color \textcolor{red}{red} indicates improvement.
\end{tablenotes}
\end{table}

\begin{table*}[]
\scriptsize
\centering
  \caption{Comparison with the SOTA methods on the nuScenes \textcolor{red}{test} set. ``C.V.", ``Motor.", ``Ped.", and ``T.C." are short for construction vehicle, motorcycle, pedestrian, and traffic cone, respectively.}
  \renewcommand\arraystretch{1}
  \resizebox{\linewidth}{!}{
  \begin{tabular}{l|c|c|c|c|c|c|c|c|c|c|c|c }
    \toprule
Method             & mAP  & NDS  & Car  & Truck & C.V. & Bus  & Trailer & Barrier & Motor. & Bike & Ped. & T.C. \\ 
\midrule
PointPillars\cite{pointpillars} & 30.5& 45.3 &68.4 &23.0& 4.1 &28.2 &23.4 &38.9 &27.4& 1.1& 59.7& 30.8\\

InfoFocus  \cite{infofocus}               & 39.5 & 39.5 & 77.9 & 31.4  & 10.7 & 44.8 & 37.3    & 47.8    & 29.0   & 6.1  & 63.4 & 46.5 \\
S2M2-SSD  \cite{s2m2}              & 62.9 & 69.3 & 86.3 & 56.0  & 26.2 & 65.4 & 59.8    & 75.1    & 61.6   & 36.4  & 84.6 & 77.7 \\
AFDetV2  \cite{afdetv2}            & 62.4 & 68.5 & 86.3 & 54.2  & 26.7 & 62.5 & 58.9    & 71.0    & 63.8   & 34.3  & 85.8 & 80.1 \\
VISTA  \cite{vista}            & 63.0 & 69.8 & 84.4 & 55.1  & 25.1 & 63.7 & 54.2    & 71.4    & 70.0   & 45.4  & 82.8 & 78.5 \\
PointPainting \cite{pointpainting}       & 46.4 & 58.1 & 77.9 & 35.8  & 15.8 & 36.2 & 37.3    & 60.2    & 41.5   & 24.1 & 73.3 & 62.4 \\
MVP  \cite{mvp}                 & 66.4 & 70.5 & 86.8 & 58.5  & 26.1 & 67.4 & 57.3    & 74.8    & 70.0   & 49.3 & 89.1 & 85.0 \\
PointAugmenting\cite{pointaugmenting} & 66.8& 71.0 &87.5 &57.3& 28.0 &65.2 &60.7 &72.6 &74.3& 50.9& 87.9& 83.6\\
Focals Conv-F\cite{focalconv} & 67.8& 71.8 &86.5 &57.5& 31.2 &68.7 &60.6 &72.3 &76.4& 52.5& 87.3& 84.6\\
VFF\cite{vff}&68.4 &72.4& 86.8& 58.1& 32.1& 70.2 &61.0 &73.9& 78.5& 52.9& 87.1& 83.8\\
UVTR\cite{uvtr} &67.1 & 71.1 & 87.5& 56.0 &33.8& 67.5& 59.5& 73.0 &73.4& 54.8 &86.3& 79.6\\
AutoAlign\cite{autoalign} & 65.8 & 70.9 & 85.9 & 55.3 & 29.6 & 67.7 & 55.6 & - & 71.5 & 51.5 & 86.4 & -\\
AutoAlignV2\cite{autoalignv2} & 68.4 & 72.4 & 87.0 & 59.0 & 33.1 & 69.3 & 59.3 & - & 72.9 & 52.1 & 87.6 & -\\
TransFusion \cite{transfusion} & 68.9 & 71.7 & 87.1 & 60.0 & 33.1 & 68.3 & 60.8 & 78.1 & 73.6 & 52.9 & 88.4 & 86.7\\
BEVFusion \cite{bevfusion-pku} & 69.2 & 71.8 & 88.1 & 60.9 & 34.4 & 69.3 & 62.1 & 78.2 & 72.2 & 52.2 & 89.2 & 85.2\\
UVTR\cite{uvtr} &67.1& 71.1& 87.5& 56.0& 33.8& 67.5& 59.5 &73.0& 73.4 &54.8 &86.3& 79.6 \\
DeepInteraction \cite{deepinteraction} &70.8& 73.4& 87.9& 60.2& 37.5& 70.8& 63.8& 80.4& 75.4& 54.5& 90.3& 87.0 \\
\midrule
CenterPoint \cite{centerpoint}                  & 58.0 & 65.5 & 84.6 & 51.0 & 17.5 & 60.2 & 53.2 & 70.9 & 53.7 & 28.7 & 83.4 & 76.7\\ 
 \rowcolor{blue!10} +our \textbf{\textcolor{black}{VoxelNextFusion}}  &66.8\textit{\fontsize{6}{0}\selectfont\textcolor{red}{+8.8}} & 69.5\textit{\fontsize{6}{0}\selectfont\textcolor{red}{+4.0}} & 85.1 & 56.6\textit{\fontsize{6}{0}\selectfont\textcolor{red}{+5.6}} & 36.7\textit{\fontsize{6}{0}\selectfont\textcolor{red}{+19.2}} & 67.3\textit{\fontsize{6}{0}\selectfont\textcolor{red}{+7.1}} & 58.6\textit{\fontsize{6}{0}\selectfont\textcolor{red}{+5.4}} & 73.3 & 77.6\textit{\fontsize{6}{0}\selectfont\textcolor{red}{+23.9}} & 45.3\textit{\fontsize{6}{0}\selectfont\textcolor{red}{+16.6}} & 83.6 & 83.4\textit{\fontsize{6}{0}\selectfont\textcolor{red}{+6.7}}  \\ 
 \midrule
VoxelNeXt\cite{voxelnext} & 64.5& 70.0 &84.6 &53.0& 28.7 &64.7 &55.8 &74.6 &73.2& 45.7& 85.8& 79.0\\
 \rowcolor{blue!10} +our \textbf{\textcolor{black}{VoxelNextFusion}}  &68.8\textit{\fontsize{6}{0}\selectfont\textcolor{red}{+4.3}} & 72.5\textit{\fontsize{6}{0}\selectfont\textcolor{red}{+2.5}} & 85.9 & 58.7\textit{\fontsize{6}{0}\selectfont\textcolor{red}{+5.7}} & 36.9\textit{\fontsize{6}{0}\selectfont\textcolor{red}{+8.2}} & 68.7\textit{\fontsize{6}{0}\selectfont\textcolor{red}{+4.0}} & 59.9\textit{\fontsize{6}{0}\selectfont\textcolor{red}{+4.1}} & 77.8 & 78.1\textit{\fontsize{6}{0}\selectfont\textcolor{red}{+4.9}} & 51.2\textit{\fontsize{6}{0}\selectfont\textcolor{red}{+5.5}} & 88.1 & 82.5\textit{\fontsize{6}{0}\selectfont\textcolor{red}{+3.5}}  \\
\bottomrule
  \end{tabular} }
  \label{tab_nuScens_test}
\begin{tablenotes}
\footnotesize
\item[1] The color \textcolor{red}{red} indicates improvement.
\end{tablenotes}
\end{table*}

\begin{table*}[tbh]
\scriptsize
\centering
\caption{Comparison with baseline on the nuScenes \textcolor{red}{validation} dataset. ‘C.V.’, ‘Ped.’, and ‘T.C.’ are short for construction vehicle, pedestrian, and traffic cone, respectively. }
\renewcommand\arraystretch{1}
\resizebox{\linewidth}{!}{
\begin{tabular}{c|c|c|c|c|c|c|c|c|c|c|c|c|c}
\hline
Dataset Split & Method & mAP & NDS & Car & Truck & C.V. & Bus & Trailer & Barrier & Motor. & Bike & Ped. & T.C. \\ 
\hline
\multirow{2}{*}{full} & CenterPoint$^*$ & 58.1 & 66.5 & 82.1 & 50.4 & 21.5 & 62.1 & 52.6 & 66.1 & 55.1 & 31.9 & 82.8 & 76.8 \\
& \cellcolor{blue!10} +our \textbf{\textcolor{black}{VoxelNextFusion}} & \cellcolor{blue!10}  67.3\textit{\fontsize{6}{0}\selectfont\textcolor{red}{+9.2}} & \cellcolor{blue!10} 70.1\textit{\fontsize{6}{0}\selectfont\textcolor{red}{+3.6}} & \cellcolor{blue!10} 83.1 & \cellcolor{blue!10} 57.2\textit{\fontsize{6}{0}\selectfont\textcolor{red}{+6.8}} & \cellcolor{blue!10} 33.1\textit{\fontsize{6}{0}\selectfont\textcolor{red}{+11.6}} & \cellcolor{blue!10} 70.1\textit{\fontsize{6}{0}\selectfont\textcolor{red}{+8.0}} & \cellcolor{blue!10} 63.8\textit{\fontsize{6}{0}\selectfont\textcolor{red}{+11.2}} & \cellcolor{blue!10} 74.1\textit{\fontsize{6}{0}\selectfont\textcolor{red}{+8.0}} & \cellcolor{blue!10} 73.0\textit{\fontsize{6}{0}\selectfont\textcolor{red}{+17.9}} & \cellcolor{blue!10} 49.9\textit{\fontsize{6}{0}\selectfont\textcolor{red}{+18.0}} & \cellcolor{blue!10} 85.2 & \cellcolor{blue!10} 83.7\textit{\fontsize{6}{0}\selectfont\textcolor{red}{+6.9}} \\  
\hline
\multirow{2}{*}{$\frac{1}{4} $} & CenterPoint$^*$ & 54.5 & 63.1 & 80.6 & 49.1 & 18.1 & 60.3 & 50.1 & 61.3 & 52.3 & 25.6 & 80.2 & 67.4 \\
& \cellcolor{blue!10}  +our \textbf{\textcolor{black}{VoxelNextFusion}} & \cellcolor{blue!10}  60.6\textit{\fontsize{6}{0}\selectfont\textcolor{red}{+6.1}} & \cellcolor{blue!10} 67.4\textit{\fontsize{6}{0}\selectfont\textcolor{red}{+4.3}} & \cellcolor{blue!10} 81.5 & \cellcolor{blue!10} 52.3\textit{\fontsize{6}{0}\selectfont\textcolor{red}{+3.2}} & \cellcolor{blue!10} 24.5\textit{\fontsize{6}{0}\selectfont\textcolor{red}{+6.4}} & \cellcolor{blue!10} 63.5\textit{\fontsize{6}{0}\selectfont\textcolor{red}{+3.2}} & \cellcolor{blue!10} 54.6\textit{\fontsize{6}{0}\selectfont\textcolor{red}{+4.5}} & \cellcolor{blue!10} 65.4 \textit{\fontsize{6}{0}\selectfont\textcolor{red}{+4.1}}& \cellcolor{blue!10} 66.1\textit{\fontsize{6}{0}\selectfont\textcolor{red}{+13.8}} & \cellcolor{blue!10} 40.6\textit{\fontsize{6}{0}\selectfont\textcolor{red}{+15.0}} & \cellcolor{blue!10} 83.4 & \cellcolor{blue!10} 74.1\textit{\fontsize{6}{0}\selectfont\textcolor{red}{+6.7}} \\ 
\hline
\multirow{2}{*}{$\frac{1}{10} $} & CenterPoint$^*$ & 47.8 & 57.3 & 79.7 & 43.7 & 13.5 & 59.5 & 23.3 & 52.2 & 46.6 & 22.4 & 79.0 & 57.8 \\
& \cellcolor{blue!10} +our \textbf{\textcolor{black}{VoxelNextFusion}} & \cellcolor{blue!10}  53.6\textit{\fontsize{6}{0}\selectfont\textcolor{red}{+5.8}} & \cellcolor{blue!10} 64.1\textit{\fontsize{6}{0}\selectfont\textcolor{red}{+5.8}} & \cellcolor{blue!10} 80.1 & \cellcolor{blue!10} 48.5\textit{\fontsize{6}{0}\selectfont\textcolor{red}{+4.8}} & \cellcolor{blue!10} 22.1\textit{\fontsize{6}{0}\selectfont\textcolor{red}{+8.6}} & \cellcolor{blue!10} 62.2\textit{\fontsize{6}{0}\selectfont\textcolor{red}{+2.7}} & \cellcolor{blue!10} 32.8\textit{\fontsize{6}{0}\selectfont\textcolor{red}{+9.5}} & \cellcolor{blue!10} 58.1\textit{\fontsize{6}{0}\selectfont\textcolor{red}{+5.9}} & \cellcolor{blue!10} 54.6\textit{\fontsize{6}{0}\selectfont\textcolor{red}{+8.0}} & \cellcolor{blue!10} 35.1\textit{\fontsize{6}{0}\selectfont\textcolor{red}{+12.7}} & \cellcolor{blue!10} 80.2 & \cellcolor{blue!10} 62.6\textit{\fontsize{6}{0}\selectfont\textcolor{red}{+4.8}} \\ 
\hline
\end{tabular}
}
\label{tab_nuScenes_val}
\begin{tablenotes}
\footnotesize
\item[1] * denotes re-implement result.
\item[2] The color \textcolor{red}{red} indicates improvement.
\end{tablenotes}
\end{table*}

\subsection{Implementation Details}

\subsubsection{Network Architecture} Since KITTI~\cite{kitti} and nuScenes~\cite{nuscenes} are distinct datasets with varying evaluation metrics and characteristics, we provide a detailed description of the \textcolor{black}{VoxelNextFusion} settings for each dataset in the following Section.

\textbf{\textcolor{black}{VoxelNextFusion} with PV-RCNN and Voxel R-CNN} We validate our \textcolor{black}{VoxelNextFusion} on KITTI~\cite{kitti} using PV-RCNN~\cite{pvrcnn} and Voxel R-CNN \cite{voxelrcnn} as the baselines. The pool radius of each level voxel features are [0.4, 0.8], [0.8, 1.2], [1.2, 2.4] and [2.4, 4.8] respectively. The input voxel size is set to (0.05m, 0.05m, 0.1m), with anchor sizes for cars at [3.9, 1.6, 1.56] and anchor rotations at [0, 1.57]. For data augmentation setting, we follow Focals Conv~\cite{focalconv}.


\textbf{\textcolor{black}{VoxelNextFusion} with CenterPoint and VoxelNeXt} We validate our \textcolor{black}{VoxelNextFusion} on the nuScenes~\cite{nuscenes} dataset using CenterPoint~\cite{centerpoint} and VoxelNeXt~\cite{voxelnext} as the baselines. The detection range for the X and Y axis is set at [-54m, 54m] and [-5m, 3m] for the Z axis. The input voxel size is set at (0.075m, 0.075m, 0.2m), and the maximum number of point clouds contained in each voxel is set to 10.


\subsubsection{Training and Testing Details}
We train VoxelFusoin with Adam optimizer and use pre-trained DeepLabv3~\cite{DeepLabV3} as our image feature extractor.
To enable effective training on KITTI~\cite{kitti} and nuScenes~\cite{nuscenes}, we utilize 8 NVIDIA RTX A6000 GPUs for network training. Specifically, for KITTI, our \textcolor{black}{VoxelNextFusion}, following our baseline~\cite{voxelrcnn, pvrcnn}, is trained 80 epochs. For nuScenes~\cite{nuscenes}, our \textcolor{black}{VoxelNextFusion}, based on~\cite{centerpoint, voxelnext}, is trained 20 epochs. 
For more details concerning our method, please refer to OpenPCDet~\cite{openpcdet}.

\subsection{Comparison with State-of-the-Arts}

\subsubsection{Performance on KITTI test set}
 As shown in Table~\ref{tab_kitti_test}, we compare \textcolor{black}{VoxelNextFusion} with the SOTA methods on KITTI test set. We note that our \textcolor{black}{VoxelNextFusion} shows outstanding performance at three difficulty levels of 3D and BEV detection (90.90\%, 82.93\%, 80.62\% in 3D APs and 94.46\%, 90.73\%, 88.34\% in BEV APs). For fair comparison, we reproduce Voxel R-CNN~\cite{voxelrcnn} and PV-RCNN~\cite{pvrcnn} as strong baselines respectively. It is worth noting that our re-implement results are almost identical to the results reported in~\cite{voxelrcnn} and ~\cite{pvrcnn}.
 Our VoxelFusoin surpasses Voxel R-CNN~\cite{voxelrcnn} on most metrics. Especially on the challenging hard level, we improve 3.2\%, 3.02\% and 1.09\% in car, pedestrian, and cyclist categories respectively.
Similarly, compared to PV-RCNN~\cite{pvrcnn}, our approach is only slightly improved on easy and moderate levels, while on hard level we surpass the baseline by a large margin. 
Compared with the multi-modal method Focals Conv~\cite{focalconv}, 
our method achieves superior performance, with improvements of 0.45\%, 0.65\%, and 3.03\% in the three levels on car AP 3D, respectively. 
Overall, our \textcolor{black}{VoxelNextFusion} performs well on the KITTI~\cite{kitti} test set. Especially on the hard level, which mostly consists of distant and small objects, this strongly demonstrates the effectiveness of our method.

\begin{table}[t]
\scriptsize
\centering
\caption{Effect of each component in our \textcolor{black}{VoxelNextFusion}. Results are reported on KITTI \textit{val} set for \textcolor{black}{car category} with \textcolor{red}{Voxel R-CNN}. "P" indicates one-to-one projection-only. \textcolor{black}{Runtime means inference time pre frame.}}

  \renewcommand\arraystretch{1}
  \resizebox{\linewidth}{!}{
  \begin{tabular}{c|c|c|c|c|c|c}
\toprule
 \multirow{3}{*}{P} & \multirow{3}{*}{P$^2$} & \multirow{3}{*}{FB} & \multicolumn{2}{c|}{Hard}                                        & \multirow{3}{*}{\#Params} & \multirow{3}{*}{Runtime} \\ \cmidrule(lr){4-5}
                                                 &                      &                                  & $AP_{3D}$(\%)           & $AP_{BEV}$(\%)                     &                                   &                       \\ 
                        \midrule
                                      &                      &                       & 82.80 & 89.06          & 7.59 M                            & 39ms                   \\
              \checkmark                         &                      &                       & 83.06\textit{\fontsize{6}{0}\selectfont\textcolor{red}{+0.26}} & 89.29\textit{\fontsize{6}{0}\selectfont\textcolor{red}{+0.23}}       & 7.74 M                           & 46ms                  \\
  \checkmark                         & \checkmark                     &                                  & 83.93\textit{\fontsize{6}{0}\selectfont\textcolor{red}{+1.13}}  &  90.06\textit{\fontsize{6}{0}\selectfont\textcolor{red}{+1.00}}  & 7.76 M                           & 49ms                  \\
\rowcolor{blue!10}     \checkmark                        & \checkmark                     & \checkmark                                & 84.59\textit{\fontsize{6}{0}\selectfont\textcolor{red}{+1.79}}  &  91.09\textit{\fontsize{6}{0}\selectfont\textcolor{red}{+2.03}}         & 7.78 M                           & 54ms                  \\ 
     \bottomrule
\end{tabular}}
\label{tab_ablation_kitti_voxel}
\begin{tablenotes}
\footnotesize
\item[1] The color \textcolor{red}{red} indicates improvement.
\end{tablenotes}
\end{table}

\begin{table}[t]
\scriptsize
\centering
\caption{Effect of each component in our \textcolor{black}{VoxelNextFusion}. Results are reported on nuScenes validation set (trained on $\frac{1}{4}$ subset) with \textcolor{red}{CenterPoint}. "P" indicates one-to-one projection-only. \textcolor{black}{Runtime means inference time pre frame.}}

  \renewcommand\arraystretch{1}
  \resizebox{\linewidth}{!}{
  \begin{tabular}{c|c|c|c|c|c|c}
\toprule
P & P$^2$ & FB & mAP  & NDS  & \#Params & Runtime \\ \midrule
        &     &      & 54.5 & 63.1 & 9.01M    & 95ms    \\
                                 \checkmark      &     &      & 56.0\textit{\fontsize{6}{0}\selectfont\textcolor{red}{+1.5}}  & 64.8\textit{\fontsize{6}{0}\selectfont\textcolor{red}{+1.7}}  & 9.16M    & 124ms    \\
                                  \checkmark     & \checkmark     &      & 58.3\textit{\fontsize{6}{0}\selectfont\textcolor{red}{+3.8}}  & 66.8\textit{\fontsize{6}{0}\selectfont\textcolor{red}{+3.7}}  & 9.19M    & 141ms    \\
                                \rowcolor{blue!10}   \checkmark    &  \checkmark    &   \checkmark    & 60.6\textit{\fontsize{6}{0}\selectfont\textcolor{red}{+6.1}}  & 67.4\textit{\fontsize{6}{0}\selectfont\textcolor{red}{+4.3}}  & 9.21M    & 151ms   \\ 
                             \bottomrule
                             
\end{tabular}}
\label{tab_ablation_nuscens}
\begin{tablenotes}
\footnotesize
\item[1] The color \textcolor{red}{red} indicates improvement. 
\end{tablenotes}
\end{table}

\begin{table}[t]
\scriptsize
\centering
\caption{Ablations on use stage and fusion scope on  KITTI \textit{val} set for \textcolor{black}{car category} with \textcolor{red}{Voxel R-CNN}.
}

  \renewcommand\arraystretch{1}
  \resizebox{\linewidth}{!}{
  \begin{tabular}{c|c|c|c|c|c|c}

\toprule
 \multirow{3}{*}{Stage}      & \multicolumn{3}{c|}{AP$_{3D}$(\%)}                                                            & \multicolumn{3}{c}{AP$_{BEV}$(\%)}                                      \\ \cmidrule(lr){2-7}     & Easy& Mod. & Hard & Easy & Mod. & Hard \\ \midrule
None                         & 92.32          & 85.06          & 82.80   & 95.48          & 91.06          & 89.06            \\
1   & \textbf{92.78}   & \textbf{86.89} & \textbf{84.59}  & \textbf{95.74}  & \textbf{92.87}  & \textbf{91.09}  \\
2                          & 92.02         & 85.65 & 82.54  & 94.43  & 91.32  & 90.32  \\
3                         & 90.89          & 84.34 & 81.00  & 92.71  & 89.21  & 88.32  \\
4                          & 88.75          & 81.87 & 77.65  & 90.82  & 87.22  & 85.31  \\
 \midrule
                             
\end{tabular}}
\label{tab_ablation_KITTI_stage}
\end{table}

\begin{table}[t]
\scriptsize
\centering
\caption{Effect of the number of $K_{off}$ on  KITTI \textit{val} set for \textcolor{black}{car category} with \textcolor{red}{Voxel R-CNN}.}

  \renewcommand\arraystretch{1}
  \resizebox{\linewidth}{!}{
  \begin{tabular}{c|c|c|c|c|c|c}

\toprule
 \multirow{3}{*}{$K_{off}$}      & \multicolumn{3}{c|}{AP$_{3D}$(\%)}                                                            & \multicolumn{3}{c}{AP$_{BEV}$(\%)}                                      \\ \cmidrule(lr){2-7}     & Easy& Mod. & Hard & Easy & Mod. & Hard \\ \midrule
9   & \textbf{92.78}   & \textbf{86.89} & \textbf{84.59}  & \textbf{95.74}  & \textbf{92.87}  & \textbf{91.09}  \\
16  & 92.02   & 86.35 & 84.54  & 95.43  & 92.32  & 90.32  \\
25  & 92.56   & 86.29 & 84.11  & 94.89  & 92.01  & 90.56  \\
36  & 92.39   & 85.97 & 84.00  & 95.01  & 92.21  & 90.78  \\

 \midrule
                             
\end{tabular}}
\label{tab_ablation_KITTI_Koff}
\end{table}

\begin{table}[t]
\scriptsize
\centering
\caption{Ablations on the importance threshold $\mathcal{T}$ on KITTI \textit{val} set for \textcolor{black}{car category} with \textcolor{red}{Voxel R-CNN}.}

  \renewcommand\arraystretch{1}
  \resizebox{\linewidth}{!}{
  \begin{tabular}{c|c|c|c|c|c|c}

\toprule
 \multirow{3}{*}{$\mathcal{T}$}      & \multicolumn{3}{c|}{AP$_{3D}$(\%)}                                                            & \multicolumn{3}{c}{AP$_{BEV}$(\%)}                                      \\ \cmidrule(lr){2-7}     & Easy& Mod. & Hard & Easy & Mod. & Hard \\ \midrule
0.1 & 92.43   & 85.35 & 83.87  & 94.38  & 92.43  & 89.68  \\
0.3 & 92.46  & 86.45 & 84.04  & 94.75  & 92.51  & 90.43  \\
0.5   & \textbf{92.78}   & \textbf{86.89} & \textbf{84.59}  & \textbf{95.74}  & \textbf{92.87}  & \textbf{91.09}  \\
0.7  & 92.45   & 86.50 & 84.39  & 95.40  & 92.53  & 90.21  \\
0.9  & 92.16   & 85.18 & 84.01 & 95.12  & 92.32  & 89.69  \\

 \midrule
                             
\end{tabular}}
\label{tab_ablation_KITTI_T}
\end{table}

\subsubsection{Performance on KITTI validation dataset}
We further provide the results of the KITTI validation set to better present the detection performance of our \textcolor{black}{VoxelNextFusion}, as shown in Table~\ref{tab_kitti_val}. There are significant improvements compared to the baseline Voxel R-CNN~\cite{voxelrcnn} and PV-RCNN~\cite{pvrcnn} on moderate and hard levels. For a multi-modal 3D object detector, the dense semantic information of images can not be fully utilized, thus limiting the performance of detection methods. 
The key factor of the effectiveness of \textcolor{black}{VoxelNextFusion} is that it can incorporate key semantic information in images.

\subsubsection{Performance on nuScenes test dataset}
We also conduct experiments on larger-scale nuScenes~\cite{nuscenes} dataset using the SOTA 3D detector CenterPoint~\cite{centerpoint} and VoxelNeXt~\cite{voxelnext} as baselines to further validate the effectiveness of our \textcolor{black}{VoxelNextFusion}, as shown in Table~\ref{tab_nuScens_test}. Based on Centerpoint~\cite{centerpoint}, our \textcolor{black}{VoxelNextFusion} achieves 66.8\% mAP and 69.5\% NDS, which surpasses the baseline by 8.8\% mAP and 4.0\% NDS.
\textcolor{black}{It is worth noting that in the ``Motor" and ``C.V." categories, our method receives remarkable improvements of 23.9\% and 19.2\% in AP respectively.}
Based on the fully sparse VoxelNeXt~\cite{voxelnext}, our method can consistently \textcolor{black}{improve} the performance with improvements of 4.3\% and 2.5\% in mAP and NDS, respectively. 
It fully demonstrates the generalization and effectiveness of our method. 
\textcolor{black}{Overall, our method improves main metrics on CenterPoint~\cite{centerpoint} and VoxelNeXt~\cite{voxelnext}, resulting in improvements of 8.8\% and 4.3\% in mAP, respectively.}
Thanks to the full fusion of image features in our P$^{2}$-Fusion, it allows significant performance improvements for small objects like ``Motor.", ``Bike", ``Ped." and ``T.C.".

\subsubsection{Performance on nuScenes validation dataset}
To demonstrate the effectiveness of our \textcolor{black}{VoxelNextFusion} framework, experiments are conducted on the nuScenes validation dataset using the CenterPoint~\cite{centerpoint} baseline. As shown in Table~\ref{tab_nuScenes_val}, our method outperforms CenterPoint by 11.6\%, 17.9\%, and 18.0\% on ``C.V.", ``Motor", and ``Bike" categories in the nuScenes full validation dataset, respectively. Additionally, our method shows significant improvement on ``Motor" and ``Bike" categories which contain a large number of small long-range objects across datasets of different sizes. These results further validate the effectiveness of \textcolor{black}{VoxelNextFusion} in detecting small objects at long distances.

\subsection{Ablation Study}
\subsubsection{Effect of P$^2$ and FB sub-modules }

This section discusses the results of ablation experiments conducted on the baseline detectors Voxel R-CNN\cite{voxelrcnn} and CenterPoint\cite{centerpoint} to evaluate the performance of each component in \textcolor{black}{VoxelNextFusion}. The results are reported in Table \ref{tab_ablation_kitti_voxel} and Table \ref{tab_ablation_nuscens} for KITTI and nuScenes $\frac{1}{4}$ subset, respectively.

Table \ref{tab_ablation_kitti_voxel} shows the initial AP scores for both AP$_{3D}$ and AP$_{BEV}$ on KITTI, which are 82.80\% and 89.06\%, respectively. Employing the one-to-one projection-only module to the image branch results in a minor improvement of only 0.26\% and 0.23\% for AP$_{3D}$ and AP$_{BEV}$, respectively. However, the subsequent addition of the P$^2$ and FB sub-modules leads to a significant improvement in performance on the hard level, with an increase of 1.79\% and 2.03\% for AP$_{3D}$ and AP$_{BEV}$, respectively. All improvements are acceptable with runtime and params.
Our \textcolor{black}{VoxelNextFusion} effectively bridges the resolution gap between point clouds and images, leading to this considerable enhancement.

As shown in Table \ref{tab_ablation_nuscens}, one-to-one projection (P) only weakly improves the performance. However, when P2Fusion is employed, there is an excellent performance improvement, which demonstrates that one-to-many projection (P$^{2}$) can better fuse image semantic features to enhance the 3D detector. Moreover, when integrated with FB-Fusion, the enhancement further amplifies to reach 6.1\% and 4.3\% improvement in mAP and NDS, respectively. Notably, in comparison to KITTI, our \textcolor{black}{VoxelNextFusion} produces more \textcolor{black}{remarkable} improvement on the large-scale nuScenes dataset. 
In summary, our ablation experiments show that \textcolor{black}{VoxelNextFusion} effectively enhances the performance of baseline on challenging datasets. The results emphasize the significance of addressing the resolution gap between point clouds and images and offer valuable insights for designing effective fusion strategies.

\begin{table}[]
\scriptsize
\centering
\caption{Performance on different distances. The results are evaluated with AP calculated by 40 recall positions and 0.7 IoU threshold for \textcolor{black}{car category} in the \textcolor{red}{hard} level on KITTI \textit{val} set.
}
\renewcommand\arraystretch{1}
\setlength{\tabcolsep}{0.95mm}{

\begin{tabular}{c|c|c|c|c|c|c}

\toprule
 \multirow{3}{*}{Method}      & \multicolumn{3}{c|}{AP$_{3D}$(\%)}                                                            & \multicolumn{3}{c}{AP$_{BEV}$(\%)}                                      \\ \cmidrule(lr){2-7} 
                                                                     & 0-20m & 20-40m & 40m-inf & 0-20m & 20-40m & 40m-inf \\ \midrule
    
                                      Voxel R-CNN$^*$        & 93.14 & 73.42  & 29.57   & 92.42 & 86.12  & 51.00
                                              \\
                                      +Focals Conv $^*$      & 94.25 & 77.27\textit{\fontsize{6}{0}\selectfont\textcolor{black}{+3.85}}  & 36.59\textit{\fontsize{6}{0}\selectfont\textcolor{black}{+7.02}}   & 93.00 & 89.22\textit{\fontsize{6}{0}\selectfont\textcolor{black}{+3.10}}  & 52.34\textit{\fontsize{6}{0}\selectfont\textcolor{black}{+1.34}}
                                              \\
                                    \rowcolor{blue!10}  +\textcolor{black}{VoxelNextFusion}      & 96.13 & 82.44\textit{\fontsize{6}{0}\selectfont\textcolor{red}{+9.02}}  & 44.49\textit{\fontsize{6}{0}\selectfont\textcolor{red}{+14.92}}   & 96.47 & 91.45\textit{\fontsize{6}{0}\selectfont\textcolor{red}{+5.33}}  & 56.56\textit{\fontsize{6}{0}\selectfont\textcolor{red}{+5.56}}
                                              \\
 \midrule
PV-RCNN$^*$        & 93.11 & 71.02  & 34.12   & 94.28 & 85.71  & 49.21
                                              \\
                                      +Focals Conv $^*$      & 93.92 & 75.12\textit{\fontsize{6}{0}\selectfont\textcolor{black}{+4.10}}  & 38.94\textit{\fontsize{6}{0}\selectfont\textcolor{black}{+4.82}}   & 96.86 & 87.65\textit{\fontsize{6}{0}\selectfont\textcolor{black}{+1.94}}  & 52.30\textit{\fontsize{6}{0}\selectfont\textcolor{black}{+3.09}}
                                              \\
                                   \rowcolor{blue!10}     +\textcolor{black}{VoxelNextFusion}      & 94.32 & 80.98\textit{\fontsize{6}{0}\selectfont\textcolor{red}{+9.96}}  & 45.98\textit{\fontsize{6}{0}\selectfont\textcolor{red}{+11.86}}   & 96.32 & 89.98\textit{\fontsize{6}{0}\selectfont\textcolor{red}{+4.27}}  & 58.31\textit{\fontsize{6}{0}\selectfont\textcolor{red}{+9.10}}
\\
\bottomrule
\end{tabular}
\label{tab_distances}
\begin{tablenotes}
\footnotesize
\item[1] * denotes re-implement result.
\item[1] The color \textcolor{black}{blue} highlights improvement to one-to-one solution.
\item[1] The color \textcolor{red}{red} indicates improvement to our \textcolor{black}{VoxelNextFusion}.
\end{tablenotes}
}
\end{table}



\begin{figure}[htb]
\centering \includegraphics[width=1\linewidth]{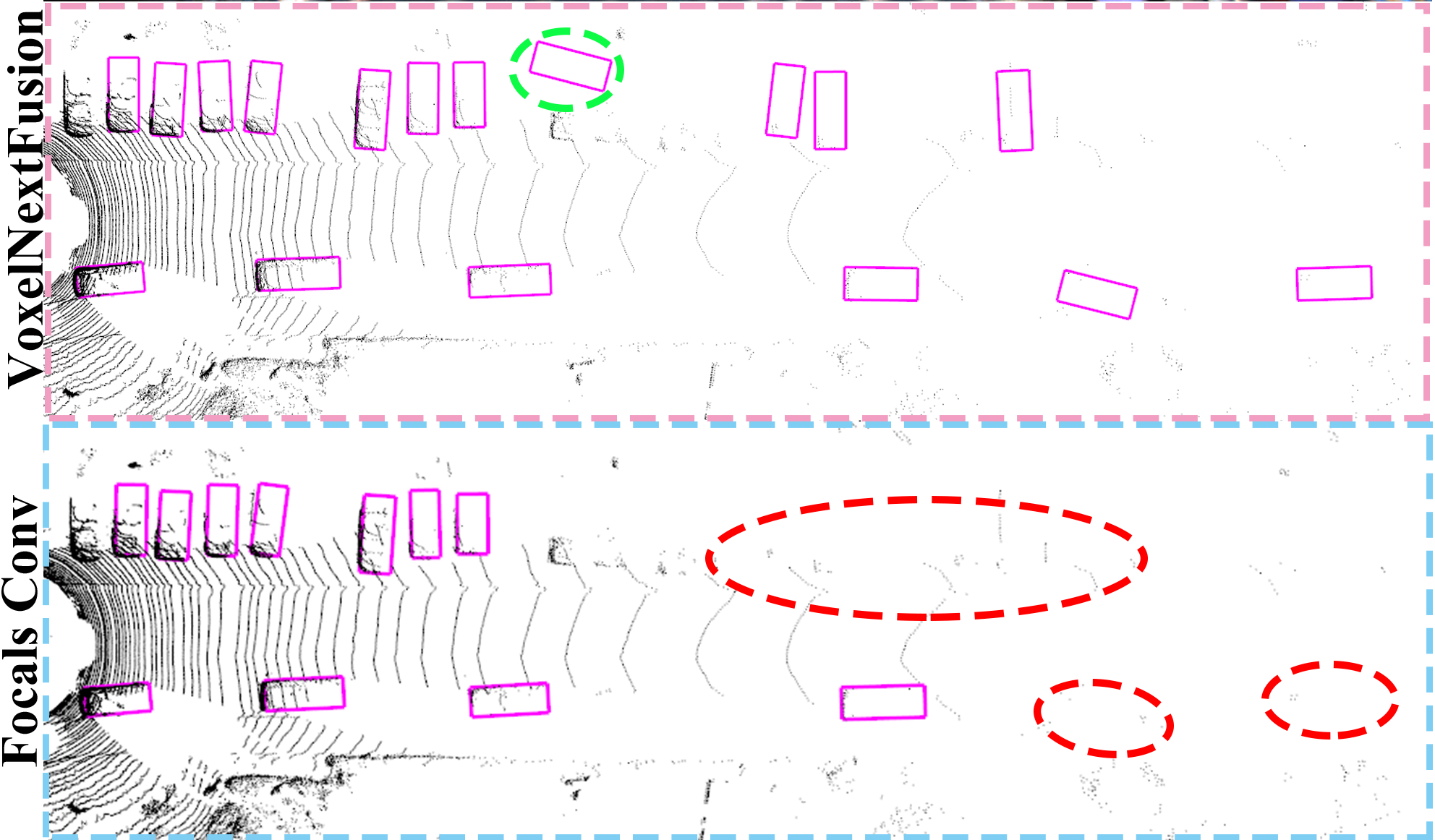}
 \caption{Visualize the comparison between Focals Conv and our \textcolor{black}{VoxelNextFusion} in long-range detection, highlighting false positives in \textcolor{green}{green} and false negatives in \textcolor{red}{red}. 
 }
\label{fig:Visualization}
\end{figure}

\subsubsection{Use Stage and Fusion Scope Analysis}
\textcolor{black}{Following baseline~\cite{voxelrcnn}, our 3D backbone consists of 4 stages to extract different scale features.} As shown in Table \ref{tab_ablation_KITTI_stage}, we validated the performance impact of \textcolor{black}{VoxelNextFusion} application at different stages of backbone on nuScenes~\cite{nuscenes}. 
We found that applying \textcolor{black}{VoxelNextFusion} in the early stages can achieve the best performance, but as the fusion stage is delayed, the performance continues to decline. This is because the feature map of early stage has a higher resolution, and image features do not require additional downsampling operations when fused with voxel features, thereby preserving more semantic information and contributing to performance improvement. 

\subsubsection{Effect of the number of $K_{off}$}
\textcolor{black}{Since the Patch feature in the $P^2$-Fusion module is a critical component of this paper, we are discussing the size and corresponding effectiveness of the Patch feature. The Patch feature is determined by the hyperparameter $K_{off}$, which serves as the neighboring pixel coordinates. Here, we have configured different values for the hyperparameter: 9, 16, 25, and 36, corresponding to the configurations [-1,0,1]$^2$, [-1,0,1,2]$^2$, [-2,-1,0,1,2]$^2$, and [-2,-1,0,1,2,3]$^2$.
As depicted in Table \ref{tab_ablation_KITTI_Koff}, the variations in $K_{off}$ do not exhibit significant impact on the performance. Notably, when $K_{off}$ is set to 9, our \textcolor{black}{VoxelNextFusion}  achieves superior performance.}

\subsubsection{Ablations on the importance threshold $\mathcal{T}$}
\textcolor{black}{
As shown in Table \ref{tab_ablation_KITTI_T}, we conducted an ablation study on the crucial threshold $\mathcal{T}$ on the KITTI validation set. The range of $\mathcal{T}$ ranged from 0.1 to 0.9. Overall, when $\mathcal{T}$ is 0.5, our VoxelNextFusion achieved the better performance, and the performance variations were not substantial. It indicating that our VoxelNextFusion is not highly sensitive to hyperparameters.
}

\subsubsection{Distances Analysis}
To better understand the superior performance of our \textcolor{black}{VoxelNextFusion} at long distances, we provide performance metrics for different distance ranges in Table \ref{tab_distances}, particularly as hard level includes more small and occluded objects. Specifically, compared to the Focals Conv\cite{focalconv} with one-to-one projection, our metrics show a more significant improvement, especially in the distance ranges of 20-40m and 40m-inf. For example, in 3D detection at 40m-inf, adding the Focals Conv improved the baseline Voxel R-CNN by only 7.02\%, while our \textcolor{black}{VoxelNextFusion} improved it by 14.92\%. In BEV detection at 40m-inf, adding Focals Conv only improved the baseline by 1.34\%, while our \textcolor{black}{VoxelNextFusion} improved it by 5.56\%. These results clearly reflect the advantages of our \textcolor{black}{VoxelNextFusion} at longer distances, primarily addressing the problem of sparse point clouds at such distances and introducing more appropriate pixel features to significantly improve the accuracy of distant objects. 

\subsection{Visualization}
In Fig.~\ref{fig:Visualization}, we illustrate the superiority of our \textcolor{black}{VoxelNextFusion} over the one-to-one projection-based approach Focals Conv for long-range object detection, while both of them utilized Voxel R-CNN\cite{voxelrcnn} as the baseline. While our \textcolor{black}{VoxelNextFusion} has a false detections, there are no instances of missed detections, whereas Focals Conv\cite{focalconv} suffers from numerous false negatives. This can be attributed to the fact that our \textcolor{black}{VoxelNextFusion} makes more reasonable use of semantic information in the image domain, without compromising on its advantages of semantic and geometric continuity, which are often crucial for exploiting the benefits of imaging in the context of long-range, sparse point clouds where geometric relationships are difficult to establish. Overall, our method exhibits significant improvement in the precision of remote object detection.

\section{Conclusions}

In this work, we propose \textcolor{black}{VoxelNextFusion}, a simple, unified, and effective voxel fusion framework for multi-modal 3D object detection. Specifically, we design a unified multi-modal framework based on four classic voxel-based approaches, Voxel R-CNN, PV-RCNN, CenterPoint, and VoxelNeXt, which makes more reasonable use of image semantic information and background information, thereby enhancing generalization and robustness. Comprehensive experimental results demonstrate that \textcolor{black}{VoxelNextFusion} significantly improves the performance of 3D detectors on the KITTI and nuScenes datasets. We hope our work can provide new insights into multi-modal feature fusion for autonomous driving.

\addtolength{\textheight}{-0cm}



\bibliographystyle{IEEEtran}
\bibliography{IEEEabrv, main}
%
%
%

\begin{IEEEbiography}[{\includegraphics[width=1in,height=1.25in,clip,keepaspectratio]{{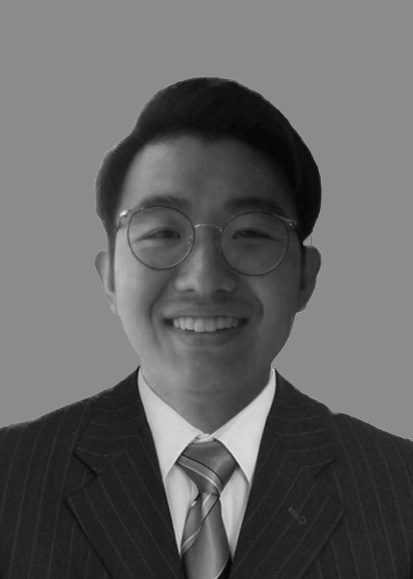}}}]{Ziying Song} was born in Xingtai, Hebei Province, China, in 1997. He received his B.S. degree from Hebei Normal University of Science and Technology (China) in 2019. He received a master's degree from Hebei University of Science and Technology (China) in 2022. He is now a Ph.D. student majoring in Computer Science and Technology at Beijing Jiaotong University (China), with research focus on Computer Vision.

\end{IEEEbiography}

\begin{IEEEbiography}[{\includegraphics[width=1in,height=1.25in,clip,keepaspectratio]{{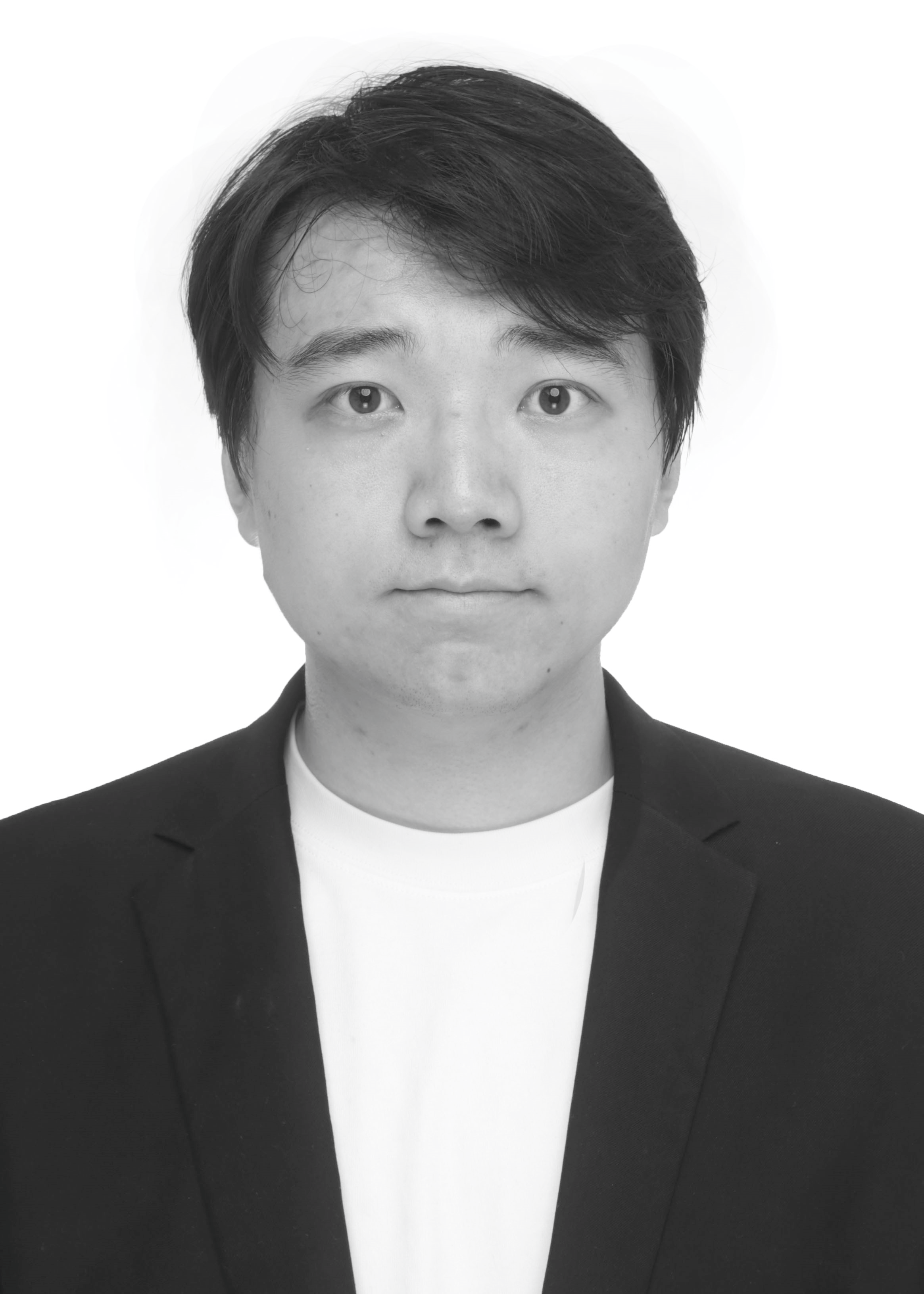}}}]{Guoxin Zhang} was born in 1998 in Xingtai, Hebei Province, China. In 2021, he received his bachelor's degree from Hebei University of Science and Technology (China). He is now studying for his master's degree at the Hebei University of Science and Technology (China). His research interests are in computer vision.

\end{IEEEbiography}

\begin{IEEEbiography}[{\includegraphics[width=1in,height=1.25in,clip,keepaspectratio]{{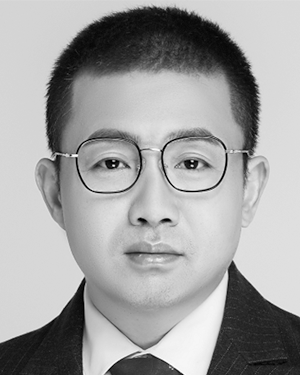}}}]{Jun Xie} was born in Zhengzhou, Henan Province, China, in 1978. He received his M.S. of EECS in 2002 University of Science and Technology of China (Beijing). Since December 2013, he has worked as Advanced Researcher of Lenovo Research. His research interests are in Computer Vision. 

\end{IEEEbiography}

\begin{IEEEbiography}[{\includegraphics[width=1in,height=1.25in,clip,keepaspectratio]{{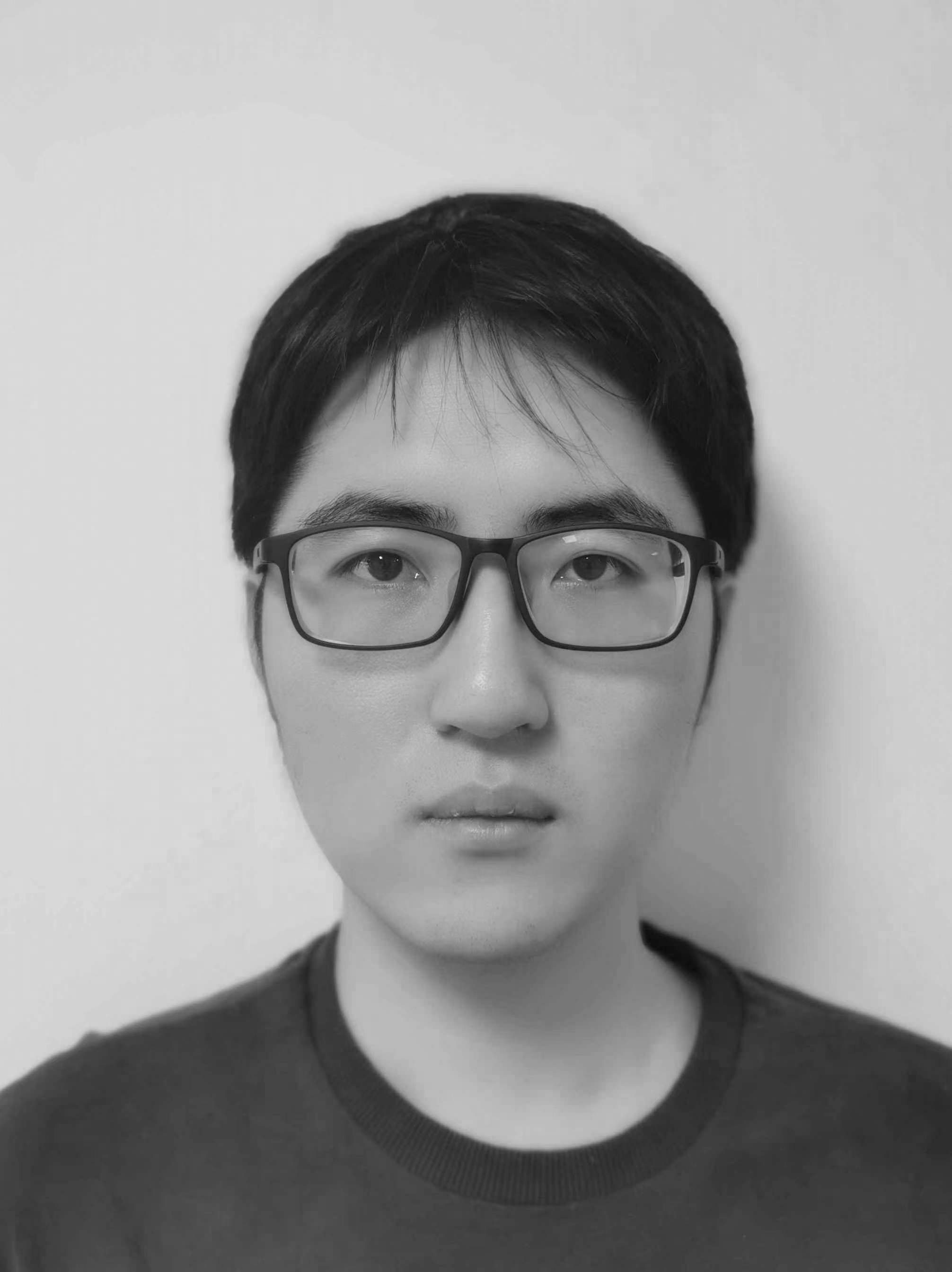}}}]{Lin Liu} was born in Jinzhou, Liaoning Province, China, in 2001. he received his bachelor’s degree from China University of Geosciences(Beijing). Now, he is studying for his master's degree at the Beijing Jiaotong University (China). His research interests are in computer vision.

\end{IEEEbiography}

\begin{IEEEbiography}[{\includegraphics[width=1in,height=1.25in,clip,keepaspectratio]{{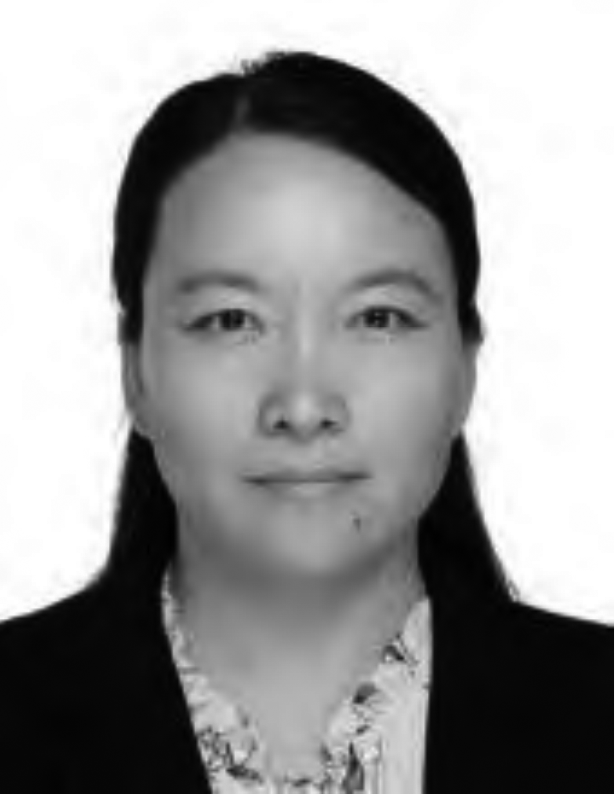}}}]{Caiyan Jia} was born in 1976. She received her Ph.D. degree from Institute of Computing Technology, Chinese Academy of Sciences, China, in 2004. She had been a postdoctor in Shanghai Key Lab of Intelligent Information Processing, Fudan University, Shanghai, China, in 2004–2007. She is now a professor in School of Computer and Information Technology, Beijing Jiaotong University, Beijing, China. Her current research interests include deep learning in computer vision, graph neural networks and social computing, etc.

\end{IEEEbiography}

\begin{IEEEbiography}[{\includegraphics[width=1in,height=1.25in,clip,keepaspectratio]{{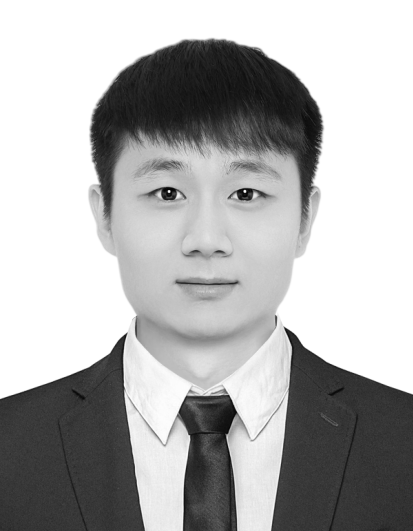}}}]{Shaoqing Xu} received his M.S. degree in transportation engineering from the School of Transportation Science
and Engineering in Beihang University. He is currently working toward the Ph.D. degree in electromechanical engineering with the State Key Laboratory of Internet of Things for Smart City, University of Macau, Macao SAR, China. His research interests include intelligent transportation systems, Robotics and computer vision.
\end{IEEEbiography}

\begin{IEEEbiography}[{\includegraphics[width=1in,height=1.25in,clip,keepaspectratio]{{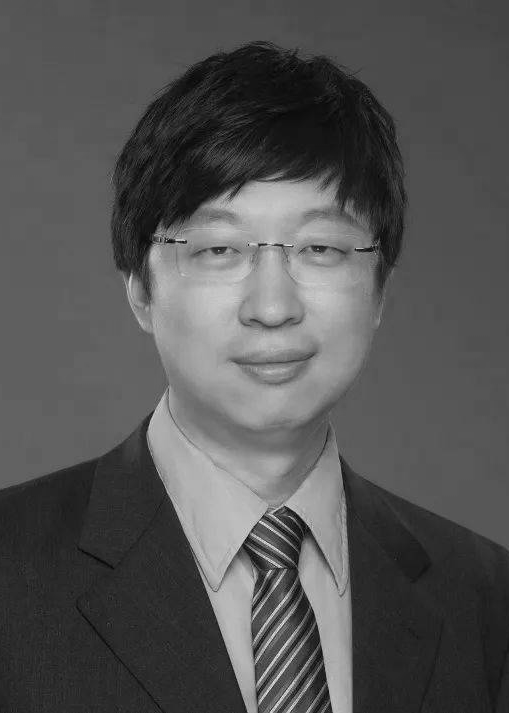}}}]{Zhepeng Wang}  was born in Yuncheng, Shanxi Province, China, in 1976. He received his B.S of EECS in 1997 and M.S. of EECS in 2000 from Tsinghua University(China). Currently, he is the Vice President of Lenovo and the Head of PC Innovation and Ecosystem Lab of Lenovo Research Institute.

\end{IEEEbiography}

\end{document}